\PassOptionsToPackage{numbers,sort&compress}{natbib}
\documentclass{ceurart}

\usepackage[subtle,tracking=normal,paragraphs=normal]{savetrees} 

\usepackage{wrapfig}
\usepackage{needspace}
\usepackage{placeins}
\usepackage{textcomp}
\usepackage{url}
\usepackage{nicefrac}
\usepackage{bm}
\usepackage{subcaption}
\usepackage{enumitem}
\usepackage{listings}
\microtypesetup{expansion=false}
\usepackage{tikz}
\usetikzlibrary{arrows.meta,positioning,fit,backgrounds,calc}
\usepackage[most]{tcolorbox}

\definecolor{amBlue}{HTML}{1F4E79}    
\definecolor{amBlueL}{HTML}{E8F0FB}   
\definecolor{amTeal}{HTML}{0F7B7B}    
\definecolor{amTealL}{HTML}{E4F4F3}   
\definecolor{amAmber}{HTML}{B26A00}   
\definecolor{amAmberL}{HTML}{FBF0DD}  
\definecolor{amGreen}{HTML}{2E7D32}   
\definecolor{amGreenL}{HTML}{E7F3E8}  
\definecolor{amGray}{HTML}{4A4A4A}

\definecolor{lightblue}{rgb}{0.9, 0.95, 1.0}
\definecolor{mygreen}{rgb}{0.01, 0.5, 0.01}
\definecolor{myred}{rgb}{0.8, 0.01, 0.01}
\definecolor{customgray}{rgb}{0.25,0.25,0.25}
\definecolor{customred}{rgb}{0.8,0.05,0.05}

\begin{document}

\copyrightyear{2026}
\copyrightclause{Copyright for this paper by its authors. Use permitted under Creative Commons License Attribution 4.0 International (CC BY 4.0).}
\conference{6th International Workshop on Scientific Knowledge: Representation, Discovery, and Assessment, Oct 2026, Bari, Italy}

\title{ArticleMiner: Ontology-Guided Knowledge Graph Construction from Scientific Publications}

\author[1,2]{Md Abrar Jahin}[%
orcid=0000-0002-1623-3859,
email=jahin@isi.edu,
url=https://abrar2652.github.io/,
]
\cormark[1]

\author[1,2]{Craig A. Knoblock}[%
orcid=0000-0002-6371-4807,
email=knoblock@isi.edu,
url=https://www.isi.edu/people-knoblock/,
]

\author[1,2]{Jay Pujara}[%
orcid=0000-0001-6921-1744,
email=jpujara@isi.edu,
url=https://www.jaypujara.org/,
]

\address[1]{USC Information Sciences Institute, Marina del Rey, CA 90292, USA}
\address[2]{Thomas Lord Department of Computer Science, University of Southern California, Los Angeles, CA 90089, USA}

\cortext[1]{Corresponding author.}

\begin{abstract}[Abstract]
Scientific papers keep much of their quantitative content in tables and supplementary files, where a number means something only through its header, caption, unit, analytical method, and the conventions of its field. Recovering the rows and columns of a table is therefore not the same as recovering the scientific fact it reports. Most semantic table-interpretation methods assume that a clean table is already available and subsequently map its cells or columns to ontology terms, whereas most publication-level extraction systems are designed for a single domain. We study a middle path: a shared process that reads a paper and its supplementary files, gathers evidence from several parsers and a language model, and reconciles that evidence, while a bounded human-authored task module for each task supplies the domain meaning. The module lists the canonical names the graph may use, the surface forms that map to them, a small set of derivation rules and validity constraints, an identity key, and the bindings used to write RDF. It defines what a task is allowed to emit; it does not try to list every convention of a field. We build four such modules (for drug-discovery chemistry, materials science, machine learning, and mineral geochemistry) in the \textsc{ArticleMiner} framework, and evaluate them on 163 papers, including a new geochemistry benchmark with expert-curated ground truth. In comparisons against a same-LLM few-shot baseline, the point estimates favor \textsc{ArticleMiner} on all four tasks, with uncertainty on the two smaller benchmarks. The geochemistry comparison also includes access to supplementary files, so its improvement cannot be attributed to domain guidance alone.
\end{abstract}


\maketitle

\section{Introduction}\label{sec:intro}
Scientific publications contain quantitative observations in tables, captions, footnotes, and supplementary spreadsheets. Reusing these observations in knowledge graphs (KGs), cross-paper search, and meta-analysis requires scientific interpretation as well as layout recovery. For example, \texttt{<0.5} under an Au (ppm) header denotes a below-detection observation, distinct from a measured value. A composition percentage may denote wt\% or mol\%, with its meaning supplied by a header, caption, or footnote. Extraction must preserve these contextual distinctions when constructing structured records.

Prior work addresses parts of this problem under different assumptions. Semantic table interpretation and table-to-KG systems map cells, columns, or rows to entities, classes, or properties in a KG or ontology~\cite{huynh2022dagobah,vu2024sand,ritze2015dbpedia,bai2024schema}, but generally assume a usable table representation is available. Scientific information extraction systems operate closer to publications but are typically engineered for a single domain or target relation, e.g., chemical properties, glass compositions, or mineral-resource records~\cite{swain2016chemdataextractor,gupta2023discomat,knoblock2026minmod}. We study an explicit task-module interface for reusing publication-level evidence acquisition and extraction across domains, while making the required domain-specific artifacts visible.

We investigate three research questions. \textbf{RQ1:} How does domain guidance affect extraction quality under a fixed backbone? \textbf{RQ2:} How can outputs from multiple evidence sources be reconciled while preserving their provenance? \textbf{RQ3:} How do acquisition failures and interpretation errors contribute to the remaining error? Same-LLM comparisons and component ablations address RQ1; the reconciliation design and backend ablations address RQ2; raw-PDF versus pre-parsed comparisons and error analysis address RQ3. The full-system comparison includes several architectural differences; ontology ablations more directly assess the contribution of domain guidance.

\textsc{ArticleMiner} separates shared publication-level orchestration (\S\ref{sec:methodology}) from a bounded, human-authored ontology module supplying vocabulary, normalization and derivation rules, executable validation, identity, and RDF bindings (Figure~\ref{fig:module}). The module defines a task's permissible outputs without axiomatizing the entire field. Adapting the framework requires this module, prompts, an output schema, and, where needed, a classifier assigning records to domain categories, such as geochemical deposit types.

This paper makes three contributions. \textbf{(1)} We formulate ontology-constrained publication-to-KG construction, applying domain knowledge during extraction, normalization, and validation rather than only after table recovery, preserving unit, null-value, identifier, and taxonomy semantics (RQ1). \textbf{(2)} We define an explicit ontology-module interface and evidence model for reconciling parser outputs into validated records with source and backend-support provenance (RQ2). \textbf{(3)} We report an empirical study of four bounded task instantiations over 163 papers, with raw-PDF and pre-parsed settings, matched same-LLM baselines, parser and component ablations, cost/runtime measurements, and JSON-LD export structure (RQ3). The evaluation distinguishes full-system comparisons from component ablations and reports uncertainty at the publication level.


\section{Problem Formulation and Method}
\label{sec:methodology}
\begin{figure}[tb]
\centering
\includegraphics[width=\linewidth]{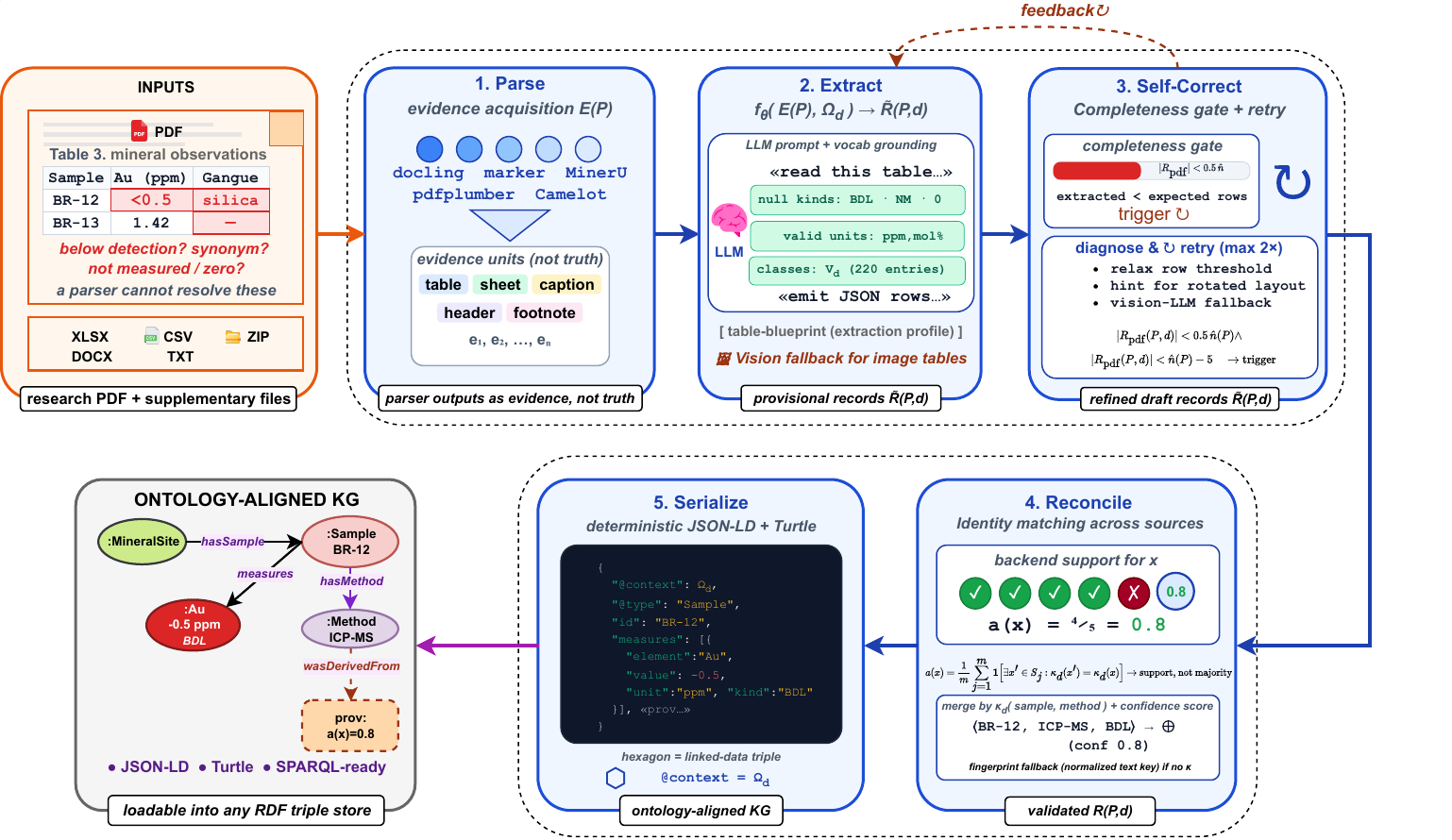}
\caption{Shared orchestration and task-module integration (schematic). Parsing preserves source locations; extraction proposes records; targeted retries address missing or unreadable evidence. Reconciliation applies derivation, canonicalization, validation, and identity-based merging. JSON-LD exposes the resulting records; Turtle and graph-query labels depict intended downstream use, not evaluated native outputs. Backend support measures identity recovery, not factual correctness.}
\label{fig:architecture}
\end{figure}

Figure~\ref{fig:architecture} summarizes the five stages: parse, extract, self-correct, reconcile, and serialize. Sections~\ref{sec:method-formalism} and \ref{sec:method-pipeline} define the shared interface and orchestration; \S\ref{sec:method-geochem-knowledge} instantiates the module for geochemistry. Each task supplies a module, prompts, a row schema, and an optional classifier. The four instantiations are GeoChem (mineral geochemistry), ChemTables (drug-discovery chemistry), DiSCoMaT (materials science), and MLTables (machine learning).

We anchor the exposition in a mineral-geochemistry example. A row reports \texttt{Au<0.5} for sample \texttt{BR-12}; the header specifies ppm, and an associated method label reads \texttt{icpms}. Recovering the observation requires three distinctions: the inequality denotes a detection limit rather than a measured concentration; the method alias maps to \texttt{ICP-MS}; and the unit comes from the header rather than the cell. This schematic example illustrates the method rather than a measured benchmark result. Each stage below shows how contextual interpretation and explicit module rules preserve this information.

\subsection{Ontology Module and Framework Operator}
\label{sec:method-formalism}
A domain module specifies the permissible outputs of a bounded extraction task. It provides canonical vocabulary, surface-form mappings, deterministic derivations, executable constraints, identity keys, and RDF bindings. Unknown terms are retained or rejected according to the task's validators; membership in the accepted-record space denotes conformance to these constraints, not independently verified factual correctness. The modules use labels, aliases, shallow hierarchies, and value conventions, without OWL reasoning. Validation runs as Python code. Simple admissibility and range checks can be represented as SHACL constraints, but no SHACL engine is executed in the evaluated pipeline.

We model a domain module as a six-tuple
\begin{equation}
\Omega_d = (V_d,\; C_d,\; \Phi_d,\; \Gamma_d,\; \kappa_d,\; \beta_d),
\label{eq:ontology}
\end{equation}
\begin{figure}[tb]
\centering
\includegraphics[width=\linewidth]{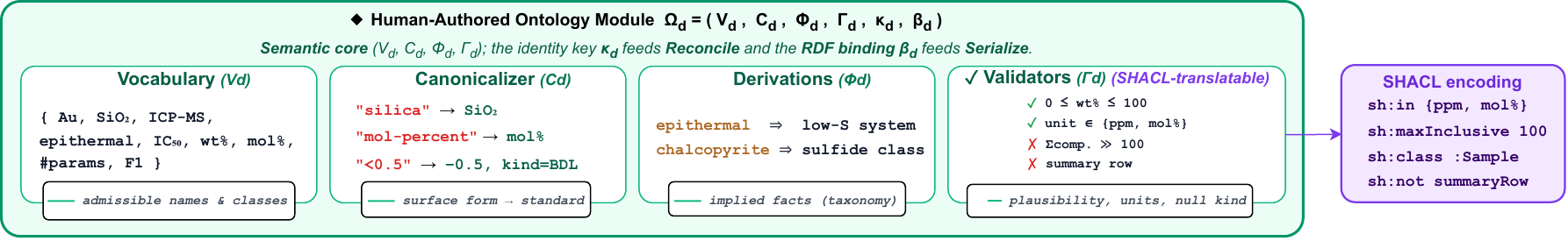}
\caption{The six-component task-module interface. The semantic core supplies vocabulary, normalization, derivations, and checks; identity and RDF bindings support reconciliation and serialization. Examples combine tasks; only declared module rules apply. Epithermal type alone does not determine sulfidation class. Validators execute as Python predicates, not as SHACL shapes.}
\label{fig:module}
\end{figure}
The component types specify the interface between provisional records, task semantics, and accepted outputs. Let \mbox{$\widetilde{\mathcal R}_d$} denote the space of \emph{provisional records}: partial assignments of the task's fields to strings or numbers, as proposed by extraction; let \mbox{$\mathcal R_d \subseteq \widetilde{\mathcal R}_d$} denote the \emph{accepted-record space} of schema-complete, canonicalized records. $V_d$ is a finite set of canonical task identifiers and types; $C_d\!:\widetilde{\mathcal R}_d\!\rightarrow\widetilde{\mathcal R}_d$ is a field-wise canonicalizer over provisional records; $\Phi_d\!:\widetilde{\mathcal R}_d\!\rightarrow\widetilde{\mathcal R}_d$ is a deterministic derivation operator; $\Gamma_d=\{\gamma_j:\widetilde{\mathcal R}_d\!\rightarrow\{0,1\}\}$ is a set of record-level validators whose conjunction gates emission; \mbox{$\kappa_d$} is the task identity key inducing an equivalence relation over accepted records (\S\ref{sec:method-pipeline}, Reconcile); and \mbox{$\beta_d$} is the RDF binding that fixes the \texttt{@context}, IRIs, and datatypes of emitted assertions (\S\ref{sec:method-pipeline}, Serialize). \mbox{$C_d$} and \mbox{$\Phi_d$} are total by construction: on surface forms outside their mapping tables, they act as the identity, so unmapped strings are preserved rather than coerced, and whether such a record is emitted is decided by \mbox{$\Gamma_d$} against the task schema. Canonical targets come from the benchmark schema and published domain resources (e.g., the USGS CMMI taxonomy and IMA mineral list for \textsc{GeoChem}); the released modules expose the curated surface-form mappings, and canonicalization precedes validation, so validators operate on canonical identifiers rather than on every synonym. Appendix~\ref{app:module-entries} gives the per-task authoring sources. The other three modules fill the same interface with different content (Figure~\ref{fig:module}, Table~\ref{tab:authoring_compact}).

Given a publication $P$ and a domain $d$, the framework produces the set of validated records
\begin{equation}
R(P,d) = \big\{ C_d(\Phi_d(r)) \,:\, r \in \tilde R(P,d),\; \Gamma_d(C_d(\Phi_d(r))) = 1 \big\},
\label{eq:rdef}
\end{equation}
where $\tilde R(P,d)\subseteq\widetilde{\mathcal R}_d$ is the set of provisional records and $\Gamma_d(q)=1$ abbreviates $\bigwedge_j\gamma_j(q)=1$. Extraction proposes; $\Phi_d$ derives declared facts; $C_d$ canonicalizes; and $\Gamma_d$ accepts or rejects. A rejected record is not emitted as an accepted record; its source locator and structured rejection reasons are retained in the audit log. Adapting to a new domain consists of authoring \mbox{$\Omega_d$} (including its identity key and RDF bindings), three to six prompt templates, the output row schema, and an optional classifier subsystem when the task contains a non-trivial classification problem; Table~\ref{tab:authoring_compact} itemizes these artifacts per task. The evaluated modules contain approximately 60--220 entries and use flat vocabularies or shallow hierarchies. Validation logs can help identify coverage gaps, although a rejection can also indicate an extraction error.

\begin{table}[tb]
\centering\footnotesize
\caption{Authored artifacts for the four tasks. Entry counts summarize module contents, not measured adaptation effort. Each task additionally needs prompts, an output schema, and RDF bindings. GeoChem uses three to six prompts and a separate deposit classifier. Detailed entries appear in Appendix~\ref{app:module-entries}.}
\label{tab:authoring_compact}
\setlength{\tabcolsep}{4pt}
\begin{tabular}{@{}>{\raggedright\arraybackslash}p{.14\linewidth}r >{\raggedright\arraybackslash}p{.37\linewidth} >{\raggedright\arraybackslash}p{.32\linewidth}@{}}
\toprule
Task & Entries & Vocabulary, rules, and checks & Identity and additional artifacts \\
\midrule
GeoChem & 220 & Elements, minerals, methods, deposit mappings; units, null conventions, hierarchy, and plausibility checks & Sample + method; 209-column projection; 189-type classifier \\
\addlinespace
DiSCoMaT & 167 & Oxide components; mol\% and wt\%; component admissibility and composition-sum checks & Material + constituent; composition tuple schema \\
\addlinespace
MLTables & 60 & Cell types, metrics, tasks, datasets; type-dependent fields and metric checks & Typed-field fingerprint; entry-type schema \\
\addlinespace
ChemTables & 80 & Assays, targets, treatments, units; value bounds and required-field checks & Treatment + target + assay; bioactivity tuple schema \\
\bottomrule
\end{tabular}
\end{table}

The module intervenes at three distinct points. Prompt-time grounding guides which fields and meanings the model proposes; deterministic derivation, canonicalization, and validation constrain the resulting records; RDF bindings specify the graph representation. These roles are complementary. A canonicalizer can normalize a supplied unit or method alias, but cannot recover a missing caption that identifies the measurement. Conversely, a model may recover all relevant context while still emitting inconsistent labels. Separating these responsibilities makes it possible to inspect whether a failure arose from absent evidence, contextual interpretation, or a declared rule, rather than treating every error as a prompting failure.

\subsection{Geochemistry Instantiation}
\label{sec:method-geochem-knowledge}
The geochemistry module illustrates four categories of semantic content within the six-component interface. Identity and RDF bindings are specified separately (Table~\ref{tab:authoring_compact}).

\emph{Vocabulary and canonicalization ($V_d,C_d$).} Mineral and analytical-method aliases map to canonical labels, such as \texttt{la-icp-ms} to \texttt{LA-ICPMS}. The approximately 220 entries summarize curated mappings and vocabulary entries; the separate deposit classifier draws on the full 189-type CMMI taxonomy. An unlisted surface form is retained for validation rather than assigned an unsupported canonical label. For example, \texttt{laser ablation icp-ms} and the misspelling \texttt{laipcms} share the canonical target \texttt{LA-ICPMS}; \texttt{electron microprobe} maps to \texttt{EPMA}. Mineral entries additionally record coarse classes and formulae, such as sphalerite, sulfide, and ZnS. The mappings encode labels, aliases, and coarse classes rather than logical axioms. A spelling absent from the curated list is not guessed: its original form is retained and assessed by the validators. These are inspectable mappings, not assertions that all scientifically related terms are interchangeable.

\emph{Taxonomic derivations ($\Phi_d$).} A recognized deposit type determines its group and environment in the adopted USGS CMMI hierarchy~\cite{hofstra2021cmio}. This is a lookup over the published classification, rather than general logical inference. For instance, the adopted taxonomy places MVT zinc-lead in the Mississippi Valley-type group and Basin hydrothermal environment. Classifying the deposit from publication evidence is a model-assisted task; filling its parent categories after that classification is deterministic. Keeping these operations separate exposes an incorrect type assignment instead of treating the derived parents as independent corroboration.

\emph{Units and null conventions.} An explicitly labeled wt\% value can be converted to ppm by multiplying by $10^4$; ambiguous units are not inferred from magnitude alone. A reported bound \texttt{<0.5} is represented as $-0.5$, preserving its detection limit. The sentinel $-99999$ denotes below detection without a stated limit, while an unmeasured value remains blank. These conventions distinguish censored measurements from measured zero. For example, a query for samples with detectable gold must exclude both kinds of below-detection observation and unmeasured cells; collapsing them into a common missing-value marker would lose that distinction. Symbols such as \texttt{n.d.} and a dash require the source legend: a task default cannot establish their meaning in every publication.

\emph{Validation ($\Gamma_d$).} Task checks examine admissible terms and units, required fields, and numerical plausibility. They constrain what is accepted, but cannot guarantee that a value is attached to the correct sample. Simple membership and range predicates have SHACL counterparts, such as \texttt{sh:in} and \texttt{sh:minInclusive}; context-sensitive extraction decisions need additional logic. A rejected candidate remains outside the accepted output, with its evidence locator and failure reason retained so that the omission can be inspected. The finite vocabulary and shallow semantics bound the scope of the module.

\subsection{Shared Orchestration}
\label{sec:method-pipeline}
The shared orchestration consumes a raw publication together with its supplementary files and produces $R(P,d)$. We describe each stage in terms of what fails without it, using the running geochemistry example as the lead illustration; the quantitative ablation evidence appears in \S\ref{sec:ablations}.

\paragraph{\textbf{Parse.}}
A single PDF backend has uneven coverage. The framework supports five PDF backends (Docling, Marker, MinerU, pdfplumber, and Camelot) and treats their outputs as evidence; the evaluation includes single-backend, leave-one-out, and all-five configurations. Docling reads publisher-rendered tables but fails on some landscape-oriented or image-only tables; pdfplumber recovers grids that other backends miss; and Camelot handles some ruled and borderless layouts but can corrupt merged headers, so no single output is trusted. Table~\ref{tab:ablation}(b) reports controlled backend configurations. MinerU lacked OCR weights in the evaluation environment, so its results do not characterize a fully enabled installation. Additional rotation handling and adaptive backend selection in the release are outside the reported evaluation (Appendix~\ref{app:repro}). An evidence unit $e\in E(P)$ records its kind (PDF table, spreadsheet sheet, caption, header, or footnote), content, source locator, and container-level annotations such as workbook and sheet names; $E(P)$ denotes the set of all evidence units associated with publication $P$.

In the running geochemistry example, raw measurements are commonly provided in multi-sheet supplementary spreadsheets. Workbook names such as \texttt{ICP-MS\_results.xlsx} may identify the analytical method, sheet names such as \texttt{sphalerite} may group observations by mineral, and column headers such as \texttt{Au\_ppm} specify the element and measurement unit. These container-level annotations are retained as evidence-unit metadata, allowing the extraction model to infer \texttt{method=ICP-MS} even when the method is not stated within an individual row.

\paragraph{\textbf{Extract.}}
An extraction call conditioned on evidence $E(P)$ and module $\Omega_d$ proposes records $f_\theta(E(P),\Omega_d)\subseteq\widetilde{\mathcal R}_d$, where $\theta$ denotes the backbone; a publication may require several calls. Prompt-time grounding supplies admissible labels, units, and null conventions. It helps the model associate a cell with the contextual fields needed by subsequent deterministic normalization. It does not replace the canonicalizer: explicitly declared aliases and value patterns remain normalizable even when returned as strings. Geochemistry additionally uses a separate 189-type deposit classifier; the other tasks omit this subsystem.

For example, the geochemistry prompt supplies method labels such as ICP-MS and XRF, unit classes such as ppm and wt\%, and the distinction between measured, below-detection, and unmeasured observations. In the running example, extraction must associate \texttt{<0.5} with Au, ppm, sample \texttt{BR-12}, and the analytical method before normalization can represent the observation correctly. The same cell string could denote a different element or unit in another table. Vocabulary constraints narrow the admissible labels, but the publication supplies the evidence connecting those labels to the cell. This is why domain guidance is applied during extraction as well as after it.

\paragraph{\textbf{Self-correct.}}
Self-correction detects likely structural failures and retries the affected acquisition stage. In the geochemistry implementation, a pre-extraction blueprint, an LLM-generated profile of the publication, estimates the expected sample count $\hat n$ from prose, captions, and table descriptions. The blueprint uses contextual descriptions to anticipate how many samples of extraction should recover. It's an estimate gate escalation rather than directly modifying measurement values; it is a diagnostic heuristic, not ground truth. When text extraction returns no samples or fewer than $0.85\hat n$ samples, an enabled vision stage reads table images and supplements the recovered sample set. After these fallbacks, a completeness retry is triggered for a nonempty PDF sample set when $|R_{\mathrm{pdf}}|<\hat n/2$ and $|R_{\mathrm{pdf}}|<\hat n-5$. These conditions require both a relative shortfall exceeding 50\% and an absolute shortfall exceeding five samples, so a small count discrepancy alone does not trigger another pass. The retry lowers the required number of recognized element columns from three to two and replaces the earlier set only if it recovers more samples.

For a failed supplementary table, a diagnostic LLM receives a preview and returns parsing hints, such as the header row, orientation, or identifier column. The parser reruns with those hints, with at most two diagnostic attempts; the PDF correction path uses one attempt. For example, identifying row~3 as the header avoids treating two rows of license text as column names. Correcting such a structural error can restore many observations at once. These retries do not establish semantic correctness: an inaccurate expected count can also encourage over-extraction. Correction and vision stages are therefore configurable, and their effects are assessed by task and backbone (\S\ref{sec:ablations}).

The diagnostic call proposes parsing parameters rather than replacement measurement values. A misplaced header, transposed table, or incorrect identifier column can suppress many rows simultaneously; correcting the responsible parameter allows the parser to recover those rows from the original evidence. The retry limit bounds repeated diagnosis, without making the resulting records inherently reliable. Vision addresses a different failure: changing text-parser parameters cannot recover cells that are absent from the machine-readable evidence.

\paragraph{\textbf{Reconcile.}}
Multiple backends and passes produce many candidates for the same underlying observation. Reconciliation first applies \mbox{$\Phi_d$} and \mbox{$C_d$} to each provisional record, validates the canonicalized record with \mbox{$\Gamma_d$}, and groups accepted candidates by the module's identity key \mbox{$\kappa_d$}, which encodes domain identity rather than string equality. In geochemistry, $\kappa_d = (\text{sample\_name}, \text{analytical\_method})$, so records for \texttt{BR-12} under \texttt{ICP-MS} from the PDF and from the supplementary spreadsheet are merged, while \texttt{ICP-MS} vs.\ \texttt{LA-ICP-MS} records for the same sample are \emph{not}. The other domains' identity keys are listed in Table~\ref{tab:authoring_compact}; machine-learning tables use a normalized text fingerprint over typed fields because no reliable semantic key exists. For each $\kappa_d$-equivalence class, let $S$ be the set of active backends and $R_s$ the candidates recovered by backend $s$. The backend support
\begin{equation}
a(x) = \frac{1}{|S|} \sum_{s \in S} \mathbb{1}\big[\exists\, x' \in R_s\colon \kappa_d(x') = \kappa_d(x)\big]
\end{equation}
records the fraction of backends recovering the same identity. It does not require agreement on every field, does not resolve conflicting values by itself, and is not a calibrated probability of correctness; backend errors can be correlated. The provenance retains source locations and backend support for inspection.
A record recovered by only one backend can still be valid; support is not a majority-vote acceptance rule. The identity key, therefore, defines which observations are candidates for reconciliation, while the remaining fields describe those observations. For example, two candidates with the same sample and method may disagree about the concentration or its unit. Their shared identity contributes to backend support, but they cannot establish which value is correct. Keeping identity, field content, and source evidence conceptually separate avoids interpreting repeated recovery of a sample name as independent verification of its measurements.

\paragraph{\textbf{Serialize.}}
The framework exposes accepted records in JSON-LD, with task-specific terms and source metadata. The RDF binding $\beta_d$ specifies how record fields map to graph predicates and datatypes. Appendix~\ref{app:jsonld} describes the released exporter\textquotesingle s record structure and its limits; serialization is not an additional evaluated prediction task. The released JSON-LD exporter and downstream RDF validation are separate from the tuple-level evaluation reported here. The exported structure separates sample identity from elemental measurements and source metadata. Sample nodes carry a paper-scoped identifier and sample name; measurement entries carry the element and value, with units when available. This structure allows consumers to retain the association between an observation and its source publication. It does not itself resolve sample identities across papers, verify external ontology mappings, or establish the correctness of a downstream query.

\subsection{A Worked Example}
\label{sec:method-walkthrough}

For sample \texttt{BR-12}, a header supplies the unit ppm and contextual evidence identifies ICP-MS as the method. Canonicalization represents \texttt{Au<0.5} as a below-detection value with a limit of 0.5, and validation checks conformance to the task schema. If four of five backends recover their identity, their support is $a(x)=0.8$; that support does not independently verify the concentration. An RDF representation can then distinguish queries for detected gold from queries for samples analyzed by ICP-MS. This is an illustrative example rather than an additional evaluated observation.

The same separation applies beyond geochemistry. In a glass-composition table, a caption declaring mol\% prevents treating a value as wt\%, even when both composition vectors sum to 100. In a bioactivity table, \texttt{IC50 >10} with unit $\mu$M is a lower bound, not a measured concentration of 10; the target and compound must remain attached to that bound. In an ML result table, \texttt{Ours: 87.3} becomes interpretable only after the caption or prose identifies the model, dataset, and metric. These examples show why output-schema conformance is necessary but insufficient: a structurally valid record can still express the wrong scientific claim. In particular, a query for samples analyzed by ICP-MS can include the running geochemistry observation, whereas a query for detected gold must exclude its censored measurement. The distinction depends on preserving the detection-limit convention together with the element and unit. Similarly, the bound bioactivity must remain attached to its compound and target; preserving the inequality alone would not identify which scientific observation it qualifies.

\section{Evaluation Design}
\label{sec:evaluation}
We evaluate \textsc{ArticleMiner} on four benchmarks (Table~\ref{tab:datasets}) chosen to span key dimensions of scientific table extraction: input modality (PDF only vs.\ PDF+supplementary), ontology size (60--220 terms), tuple granularity (cell-, row-, or sample-level), and reliance on supplementary evidence. The benchmark sizes are severely imbalanced (111 \textsc{DiSCoMaT} papers, 28 \textsc{GeoChem}, 15 \textsc{MLTables}, and 9 \textsc{ChemTables}), so the two smallest datasets are treated throughout as case evidence rather than population-level confirmation, and the 163 papers are not 163 independent draws of domain portability. Primary comparisons hold the backbone fixed and compare publication-level extraction with few-shot prompting; the pre-parsed setting, backend, and component ablations, and per-domain best configurations are secondary or post-hoc analyses and are labeled as such.

\subsection{Benchmark Datasets}
\label{sec:bench-datasets}

\textsc{GeoChem}, introduced in this paper, contains 28 peer-reviewed mineral-geochemistry papers with ground truth curated by domain experts from the U.S.\ Geological Survey. The annotations follow the \textsc{CMiO-MIN} record structure: each gold record represents a mineral observation and specifies hierarchical sample identifiers and elemental measurements, linked to mineral, analytical-method, and deposit metadata. Deposit types are drawn from the 189-category CMMI classification scheme of Hofstra et al.~\cite{hofstra2021cmio}. The benchmark contains 1{,}236 unique sample identifiers and 87{,}759 non-null elemental measurements. It is the largest benchmark in our evaluation by number of observations and the only one in which extracted records are matched to the ground truth using sample identifiers. We use the complete 28-paper corpus for all primary \textsc{GeoChem} results.

Three external benchmarks evaluate the framework in other domains. \textsc{ChemTables} contains XML-formatted drug-discovery tables; each gold record links a numeric cell to its value, measurement type (e.g., IC$_{50}$), biological target, treatment, and unit \cite{bai2024schema}. \textsc{DiSCoMaT} contains CSV-formatted glass-composition tables with gold tuples specifying the material, chemical component, percentage, and unit (mol\% or wt\%) \cite{gupta2023discomat}. \textsc{MLTables} contains {\LaTeX}-formatted machine-learning tables whose cells are labeled as \emph{Result}, \emph{Hyper-parameter}, \emph{Data Statistics}, or \emph{Other} \cite{bai2024schema}; interpreting these cells often requires evidence from the surrounding paper.

\begin{table}[ht]
\centering\small
\centering
\caption{Benchmark sizes and approximate module counts.}
\label{tab:datasets}
\setlength{\tabcolsep}{4pt}
\begin{tabular}{lcccc}
\toprule
\textbf{Dataset} & \textbf{Publications} & \textbf{Tables} & \textbf{Gold records} & \textbf{Module entries} \\
\midrule
\textsc{DiSCoMaT}  & 111 & 175 & 4{,}755 & 167 \\
\textsc{GeoChem}   &  28 & $\sim$80 & 5{,}307 & 220 \\
\textsc{MLTables}  &  15 &  68 & 2{,}060 & \phantom{0}60 \\
\textsc{ChemTables}&   9 &  14 &    462 & \phantom{0}80 \\
\bottomrule
\end{tabular}

\end{table}

\begin{table}[ht]
\centering\small
\centering
\caption{Cost (\$) and runtime (s) per paper.}
\label{tab:cost}
\setlength{\tabcolsep}{3.5pt}
\begin{tabular}{l cc cc cc}
\toprule
& \multicolumn{2}{c}{\textbf{Sonnet}} & \multicolumn{2}{c}{\textbf{Haiku}} & \multicolumn{2}{c}{\textbf{Few-shot}} \\
\cmidrule(lr){2-3}\cmidrule(lr){4-5}\cmidrule(lr){6-7}
\textbf{Task} & s & \$ & s & \$ & s & \$ \\
\midrule
\textsc{ChemTables}   &  60 & 0.10 &  86 & 0.03 & 17 & 0.05 \\
\textsc{MLTables}     & 120 & 0.10 & 118 & 0.03 & 62 & 0.05 \\
DiSCoMaT  & 140 & 0.10 &  40 & 0.03 & 25 & 0.05 \\
GeoChem   & 180 & 0.30 & 120 & 0.10 & 21 & 0.05 \\
\bottomrule
\end{tabular}

\end{table}

\subsection{Four-Tier Evaluation for \textsc{GeoChem}}
\label{sec:bench-fourtier}

A wrong deposit type, a missing sample, an incorrect concentration, and an incorrectly populated missing value affect scientific reuse differently. The four tiers separate these errors; numerical accuracy receives the largest weight because concentrations drive downstream geochemical analysis.

GeoChem uses the composite score $S=0.30T_1+0.40T_2+0.15T_3+0.15T_4$. The tiers assess different properties and must be interpreted separately. In the released evaluator, $T_1$ (metadata) averages normalized string-similarity scores for paper metadata; absent gold fields receive full credit. $T_2$ (numerical accuracy) averages numerical scores over matched rows: nonzero values within 5\% relative error receive full credit, with linearly decreasing partial credit up to 100\% error; zero gold values use an absolute tolerance of 0.001. Below-detection sentinels have separate partial-credit rules. $T_3$ (sample recovery) measures sample matching, using identifier matching with a position-based fallback when fewer than 30\% of gold rows match by name. $T_4$ (missing-value agreement) scores whether fields absent in the gold remain null; non-null gold values, including below-detection observations, are handled by $T_2$. Thus, $T_2$ and $T_4$ are conditional on row matching and do not independently measure extraction recall or full three-way null semantics. Appendix~\ref{app:schema} specifies the 209-column projection, and Appendix~\ref{app:metric-details} gives the remaining scoring details.

\subsection{Evaluation Metrics}
\label{sec:bench-metrics}
For the three external benchmarks, we report precision, recall, and strict tuple-$F_1$: a prediction counts as correct only when the required tuple fields match. GeoChem instead uses sample matching and the four-tier score defined above; its scores are not interchangeable with strict tuple-$F_1$. For \textsc{DiSCoMaT}~\cite{gupta2023discomat}, this matches the published Tuple-$F_1$ setting. Our setting is stricter than the attribute-level token-$F_1$ of Bai et al.~\cite{bai2024schema}, which gives partial credit for $\geq$25\% token overlap; their headline numbers are therefore not directly comparable to ours and appear only as context (\S\ref{sec:results-external}). For \textsc{GeoChem}, the four-tier score is primary; we also report sample-row P/R/$F_1$ for cross-benchmark comparability.

\subsection{Baselines, Configurations, and Statistical Protocol}
\label{sec:setup}

\textbf{Comparison matrix.}
We compare \textsc{ArticleMiner} against the same-LLM few-shot baselines and published task-specific systems. These are system comparisons rather than interventions on domain knowledge alone. \textbf{B1, PDF few-shot}: the same backbone receives the raw-PDF text in a single call with a fixed three-exemplar prompt, temperature zero, the task output schema, and the same metric scripts. It does not receive ontology grounding, executable validation, cross-backend reconciliation, vision, retries, or supplementary-file orchestration. In GeoChem, ArticleMiner reads supplementary spreadsheets while B1 receives PDF text and a compact row schema; both are scored on the common evaluation projection. The comparison, therefore, includes unequal evidence access, schema presentation, and identifier handling. \textbf{B2, pre-parsed few-shot}: the same protocol starts from the benchmark's released table text, separating evidence acquisition from extraction. \textbf{Published systems}: the GNN of Gupta et al.~\cite{gupta2023discomat} on \textsc{DiSCoMaT} is compared under tuple-$F_1$; InstrucTE of Bai et al.~\cite{bai2024schema} on \textsc{ChemTables}/\textsc{MLTables} uses attribute-level token-$F_1$ and is reported only as non-comparable context. All five closed backbones are evaluated on all four benchmarks; the full grid in the appendix additionally reports three open 7--8B backbones.
\\
\textbf{Hardware, software, and protocol.}
We run on a Linux workstation (Ubuntu 22.04) with one NVIDIA H100 80~GB for open-weight inference (vLLM 0.6.x); closed backbones are called via official APIs. PDF backends are docling, marker, MinerU, pdfplumber, and Camelot (version families, model identifiers, and environment notes in Appendix~\ref{app:repro}). Temperature is zero; resampling and exemplar ordering use seed 42. Table~\ref{tab:significance}(b) reports mean per-paper scores and percentile bootstrap intervals (1,000 resamples), paired by publication identifier. Sonnet costs approximately \$0.10 per external-benchmark paper and \$0.30 per GeoChem paper; the corresponding Haiku costs are \$0.03 and \$0.10 (Table~\ref{tab:cost}). A \texttt{no\_pdf} record denotes one of two \textsc{GeoChem} data-reuse entries without a standalone source PDF, not a model or pipeline failure. Such entries are excluded from PDF-dependent averages: headline spreadsheet-aware runs use 28 papers, open-LLM PDF runs use 26, and the paired analysis uses the 25 papers with valid outputs from both compared systems.


\section{Results}
\label{sec:experiments}
The section follows the research questions: the matched comparison (\S\ref{sec:results-external}, Table~\ref{tab:main_closed}) and component ablations (\S\ref{sec:ablations}, Table~\ref{tab:ablation}a) answer RQ1; the parser and backbone analyses (\S\ref{sec:sensitivity}, Table~\ref{tab:ablation}b) answer RQ2; the raw-PDF/pre-parsed contrast and error analysis (\S\ref{sec:errors}, Table~\ref{tab:error_dist}) answer RQ3.

\begin{table}[ht]
\centering
\footnotesize
\caption{Same-backbone comparison in the raw-publication setting. (a) Strict tuple-$F_1$ on the three external tasks; GeoChem reports composite $S$ and sample-$F_1$ on 28 entries. ArticleMiner also reads GeoChem supplements; few-shot uses PDF text. Bold marks the larger score within each pair. (b) Reported Sonnet per-paper means, percentile 95\% bootstrap intervals, and paired differences; GeoChem uses 25 paired entries. These historical summaries are descriptive; exact run provenance remains incomplete. Panel (b) is not obtained by subtracting panel (a)'s aggregate scores; rounding can also affect displayed differences.}
\label{tab:main_closed}
\label{tab:significance}
\setlength{\tabcolsep}{4.5pt}
\begin{tabular}{l l ccccc}
\toprule
\rowcolor{amBlueL}\multicolumn{7}{l}{\textbf{\textcolor{amBlue}{(a) Same-backbone comparison across five closed backbones}}} \\
\rowcolor{amBlueL!50}\textbf{Backbone} & \textbf{System} & \textbf{ChemTables} & \textbf{MLTables} & \textbf{DiSCoMaT} & \textbf{GeoChem $S$} & \textbf{GeoChem $F_1$} \\
\midrule
\multirow{2}{*}{Sonnet 4.6} & \textsc{ArticleMiner} & \textbf{32.1} & \textbf{52.2} & \textbf{79.3} & 76.0 & \textbf{60.8} \\
                            & Few-shot              & 15.8 & 43.2 & 64.5 & -- & \phantom{0}0.8 \\
\multirow{2}{*}{Opus 4.6}   & \textsc{ArticleMiner} & \textbf{31.8} & \textbf{55.9} & \textbf{81.0} & 76.3 & \textbf{60.6} \\
                            & Few-shot              & 15.9 & 41.6 & 59.5 & -- & \phantom{0}3.0 \\
\multirow{2}{*}{Haiku 4.5}  & \textsc{ArticleMiner} & \textbf{25.9} & \textbf{51.6} & \textbf{72.4} & 75.5 & \textbf{54.3} \\
                            & Few-shot              & 15.8 & 38.0 & 55.8 & -- & \phantom{0}4.1 \\
\multirow{2}{*}{GPT-4o}     & \textsc{ArticleMiner} & \textbf{30.2} & \textbf{45.6} & \textbf{61.9} & 75.9 & \textbf{63.1} \\
                            & Few-shot              & 16.5 & 32.8 & 51.9 & -- & \phantom{0}3.6 \\
\multirow{2}{*}{Gemini 2.5} & \textsc{ArticleMiner} & \textbf{22.6} & \textbf{42.7} & \textbf{75.9} & 75.7 & \textbf{57.6} \\
                            & Few-shot              & 14.3 & 42.4 & 52.1 & -- & \phantom{0}0.3 \\
\midrule
\rowcolor{amBlueL}\multicolumn{7}{l}{\textbf{\textcolor{amBlue}{(b) Sonnet 4.6 paired statistics} (mean [95\% CI]; $\Delta$)}} \\
& \multicolumn{2}{c}{\textbf{System mean}} & \multicolumn{2}{c}{\textbf{Baseline mean}} & \multicolumn{2}{c}{\textbf{$\Delta F_1$}} \\
\midrule
\textsc{ChemTables} ($n{=}9$) & \multicolumn{2}{c}{36.4 [10.8,\,66.6]} & \multicolumn{2}{c}{27.6 [5.9,\,54.8]} & \multicolumn{2}{c}{$+8.8$} \\
\textsc{MLTables} ($n{=}15$) & \multicolumn{2}{c}{52.2 [36.2,\,68.1]} & \multicolumn{2}{c}{37.8 [25.5,\,51.4]} & \multicolumn{2}{c}{$+14.4$} \\
DiSCoMaT ($n{=}111$) & \multicolumn{2}{c}{66.5 [58.1,\,73.8]} & \multicolumn{2}{c}{51.1 [43.3,\,59.2]} & \multicolumn{2}{c}{$+15.5$} \\
GeoChem ($n{=}25$) & \multicolumn{2}{c}{63.2 [49.9,\,76.1]} & \multicolumn{2}{c}{0.8 [0.3,\,1.6]} & \multicolumn{2}{c}{$+62.3$} \\
\bottomrule
\end{tabular}
\end{table}

\subsection{Same-Backbone Comparison and Uncertainty}
\label{sec:results-external}
In the raw-publication comparison, ArticleMiner has higher point estimates than the same-LLM few-shot baseline on all four tasks and all five closed backbones (Table~\ref{tab:main_closed}a). Panel~(b) reports publication-level means and paired differences, rather than the aggregate scores in panel~(a). Reported mean gains are 8.8 points on ChemTables, 14.4 on MLTables, 15.5 on DiSCoMaT, and 62.3 sample-$F_1$ points on the 25-paper GeoChem paired subset. The small ChemTables and MLTables samples yield substantial uncertainty. The paired summaries are retained as descriptive evidence; their configuration-specific run provenance is incomplete, so they do not establish a formal significance claim. The GeoChem comparison includes access to supplementary spreadsheets and is not an isolated test of ontology grounding.

For the five closed backbones, ArticleMiner scores higher with pre-parsed input on all three external benchmarks (Tables~\ref{tab:consolidated_all}--\ref{tab:grid_geochem}); GeoChem has no separate pre-parsed comparison in that table. For example, Opus~4.6 reaches 81.0 tuple-$F_1$ on raw DiSCoMaT PDFs and 94.7 on pre-parsed tables. This gap indicates acquisition losses, but does not isolate parsing from changes in evidence representation or prompting. The published DiSCoMaT GNN score of 70.04~\cite{gupta2023discomat} uses tuple-$F_1$ and is reported as external context, without a paired significance test. On GeoChem, Sonnet's reported tier means are $T_1=71.6$, $T_2=76.5$, $T_3=68.5$, and $T_4=90.6$, with $S=76.0$. Numerical and null scores are conditional on matched rows; they cannot establish that all values were recovered or assigned correctly. The baseline's low sample scores reflect both identifier mismatch and the lack of the supplementary file path. These differences explain why the gap from ArticleMiner cannot be attributed to ontology guidance alone.

\subsection{Contribution of the Domain Specification: Component Ablations}
\label{sec:ablations}
\begin{table}[ht]
\centering
\footnotesize
\caption{Haiku~4.5 component and backend analyses. (a) Component removals: tuple-$F_1$ on ChemTables/MLTables/DiSCoMaT ($n=9/15/111$); GeoChem reports sample-$F_1/S$. The GeoChem full row is the 28-paper headline reference, while component-removed runs were collected separately. Equal ablated scores do not imply equality with that reference. (b) Controlled backend configurations; GeoChem uses 26 common entries with corrected ground truth. Compare within each panel and run condition. Bold marks the best displayed value; $\uparrow$ indicates improvement over the component reference. Sonnet sensitivity is in Table~\ref{tab:ablation_sonnet}.}
\label{tab:ablation}
\label{tab:backend_ablation}
\setlength{\tabcolsep}{4.5pt}
\begin{tabular}{l ccc c}
\toprule
\rowcolor{amTealL}\multicolumn{5}{l}{\textbf{\textcolor{amTeal}{(a) Component ablation}}} \\
\rowcolor{amTealL!50}\textbf{Variant} & \textbf{ChemTables} & \textbf{MLTables} & \textbf{DiSCoMaT} & \textbf{GeoChem $F_1$\,/\,$S$} \\
\midrule
Full pipeline    & \textbf{25.9} & \textbf{51.6} & 72.4 & \textbf{54.3\,/\,75.5} \\
w/o ontology     & 23.7 & 43.9 & 56.1 & 51.0\,/\,73.2 \\
w/o self-correct & 23.1 & 49.8 & 67.1 & 51.0\,/\,74.5 \\
w/o validation   & 21.7 & 41.3 & 71.2 & 51.0\,/\,74.4 \\
w/o intelligence & 23.4 & 46.3 & \textbf{76.5}$^{\uparrow}$ & 51.0\,/\,72.7 \\
w/o vision       & 22.3 & 48.4 & 71.6 & 51.0\,/\,74.2 \\
\midrule
\rowcolor{amTealL}\multicolumn{5}{l}{\textbf{\textcolor{amTeal}{(b) PDF-backend configurations ($F_1$)}}} \\
\rowcolor{amTealL!50}\textbf{Configuration} & \textbf{ChemTables} & \textbf{MLTables} & \textbf{DiSCoMaT} & \textbf{GeoChem} \\
\midrule
docling only        & \textbf{26.2} & \textbf{56.3} & 71.7 & 47.9 \\
marker only         & 23.2 & 51.4 & 66.2 & 48.0 \\
mineru only         & 23.2 & 51.6 & 66.5 & 46.3 \\
pdfplumber only     & 23.4 & 51.0 & 66.3 & 46.3 \\
camelot only        & 25.5 & 43.2 & 43.7 & 48.0 \\
leave-one-out\ docling     & 24.2 & 41.9 & 43.5 & 46.0 \\
leave-one-out\ camelot     & 24.8 & 55.3 & \textbf{72.5} & 47.9 \\
all-five fusion     & 20.8 & 47.6 & 71.0 & \textbf{54.3} \\
\bottomrule
\end{tabular}
\end{table}

The full ablation grid uses Haiku~4.5 to limit inference cost; Sonnet~4.6 reruns assess backbone sensitivity.
Under Haiku~4.5, removing ontology grounding reduces tuple-$F_1$ by 16.3 points on DiSCoMaT, 7.7 on MLTables, and 2.2 on ChemTables (Table~\ref{tab:ablation}a). GeoChem's full row reports $54.3/75.5$ for sample-$F_1/S$; every component-removed row reports $51.0$ sample-$F_1$, while $S$ ranges from 72.7 to 74.5. Equality among the ablated rows does not imply equality with the full pipeline. Because these rows combine a headline reference with separately collected component runs, their GeoChem differences require configuration and ground-truth alignment before a causal interpretation.

Component effects also depend on the backbone. On Sonnet~4.6, removing vision reduces ChemTables from 32.1 to 14.6 and MLTables from 52.2 to 4.8; on Haiku, validation removal causes larger losses than vision removal on both tasks. Removing paper intelligence raises ChemTables from 32.1 to 44.6 under Sonnet but lowers it under Haiku (25.9 to 23.4). On Haiku DiSCoMaT, its removal raises $F_1$ from 72.4 to 76.5. These are diagnostic results, not evidence for a universally optimal component set; Table~\ref{tab:ablation_sonnet} gives the Sonnet details. Thus, a richer pipeline is not automatically a better extractor. A blueprint or retry can expose missing evidence, but can also introduce additional candidates that are irrelevant or incorrectly interpreted. The observed reversals motivate evaluating correction policies jointly with the task and backbone. They do not establish which intermediate error caused a particular gain or loss.

\subsection{Acquisition, Backends, and Backbone Sensitivity}
\label{sec:sensitivity}

\textit{PDF backends.} On \textsc{DiSCoMaT}, \texttt{docling} alone reaches $71.7$ $F_1$; the leave-Docling-out configuration scores $43.5$, compared with $71.0$ for all-five fusion. On \textsc{ChemTables} and \textsc{MLTables}, single \texttt{docling} exceeds all-five fusion ($26.2$ vs.\ $20.8$; $56.3$ vs.\ $47.6$), a negative result for universal multi-backend fusion. On \textsc{GeoChem}, all single and leave-one-out variants cluster at $46$--$48$~$F_1$, while all-five fusion reaches $54.3$ on the same papers (Table~\ref{tab:ablation}(b)). The number of backends is therefore a per-task choice rather than a fixed prescription.
\textit{LLM backbones.} On \textsc{GeoChem}, closed backbones cluster tightly at $75.5$--$76.3$ Overall $S$ while open $7$--$8$B models trail at $52.6$--$54.0$, under a different 26-paper subset and a 512-token stage cap, with deposit-classification failures reported in Appendix~\ref{app:open-geochem}. This comparison does not isolate model capacity.
\textit{Few-shot count.} The ChemTables exemplar-count analysis plateaus near three examples (Figure~\ref{fig:shot_scaling}); it does not establish that context budgets are sufficient for all tasks. Cross-LLM agreement is $52$--$58\%$ on the external benchmarks but $75\%$ on \textsc{GeoChem} (Table~\ref{tab:cross_llm_agreement}, Appendix~\ref{app:results}).

\begin{table}[ht]
\centering
\scriptsize
\caption{Error distribution (\% of error events; $n_{\text{err}}$ = count) for \textsc{ArticleMiner} (Sonnet~4.6) vs.\ the same-LLM few-shot baseline. Classes: over-extraction, omission, attribute mismatch, near miss. GPT-4o rows in Appendix~\ref{app:results}.}
\label{tab:error_dist}
\setlength{\tabcolsep}{5pt}
\begin{tabular}{ll cccc c}
\toprule
\textbf{Task} & \textbf{System} & \textbf{Over-extraction} & \textbf{Omission} & \textbf{Attribute} & \textbf{Near} & $n_{\text{err}}$ \\
\midrule
\textsc{ChemTables}   & \textsc{ArticleMiner} & 57.0 & 16.1 & 26.6 & 0.4 &  704 \\
          & Few-shot              & 10.5 & 70.9 & 18.0 & 0.7 &  440 \\
\textsc{MLTables}     & \textsc{ArticleMiner} & 59.9 & 30.7 &  3.9 & 5.5 & 2163 \\
          & Few-shot              & 27.6 & 49.8 &  5.8 & 16.7 & 1679 \\
DiSCoMaT  & \textsc{ArticleMiner} &  2.7 & 92.1 &  5.2 & 0.0 & 1721 \\
          & Few-shot              & 12.1 & 75.5 &  6.5 & 2.5 & 2697 \\
\bottomrule
\end{tabular}
\end{table}

\subsection{Error Analysis}\label{sec:errors}
\looseness-1{} The diagnostic taxonomy separates unmatched predictions (over-extraction), unrecovered gold values (omission), matched values with incorrect type/component or unit (attribute mismatch), and otherwise unmatched numerical values within 5\% of a gold value (near miss). These categories describe value-level matching and do not replace strict tuple evaluation. Errors fall into these four classes (Table~\ref{tab:error_dist}); model choice mostly shifts the precision--recall balance (Fig.~\ref{fig:pr_scatter}). Sonnet over-extracts on \textsc{ChemTables}/\textsc{MLTables} but under-extracts on \textsc{DiSCoMaT}; GPT-4o under-extracts everywhere, with omission $59$--$83\%$ of its errors (Table~\ref{tab:error_gpt4o}, Appendix~\ref{app:results}); omission accounts for approximately 50--75\% of the baseline error events, not of all gold tuples. Open-weight models on \textsc{DiSCoMaT} show high precision but low recall (P=79--88\%, R=10--43\%): validation filters a strong generator's invalid tuples but cannot recover ones never proposed, showing that schema validation alone cannot recover omitted candidates.

The error counts and proportions must be read together. For Sonnet on DiSCoMaT, omissions constitute 92.1\% of ArticleMiner's 1,721 error events, compared with 75.5\% of 2,697 for few-shot prompting. The larger omission share, therefore, does not imply more omissions in absolute terms. On ChemTables and MLTables, over-extraction instead contributes 57.0\% and 59.9\% of ArticleMiner's errors. These contrasting profiles motivate different interventions: inspect missing evidence in the former setting and candidate relevance in the latter, rather than applying uniformly stricter validation.



\section{Related Work}
\label{sec:related}
\textit{\textbf{Table-to-KG and semantic table annotation.}}\looseness-1{} The SemTab challenge and systems such as DAGOBAH and SAND, with domain-independent interpretation and DBpedia table matching, annotate cells, columns, and properties over already-recovered tables~\cite{jimenezruiz2020semtab,huynh2022dagobah,vu2024sand,vu2025semantic,ritze2015dbpedia}. Because a cell's meaning depends on its surrounding evidence (headers, captions, units, sheet names, null conventions), \textsc{ArticleMiner} instead couples parsing with interpretation.
\\
\textit{\textbf{Scientific IE and domain-specific KGs.}}\looseness-1{} ChemDataExtractor, DiSCoMaT, and MinMod~\cite{swain2016chemdataextractor,gupta2023discomat,knoblock2026minmod} show the value of domain knowledge but are each engineered around one domain, source, or target schema. \textsc{ArticleMiner} instead externalizes domain knowledge into a module over a shared orchestration, so adaptation centers on explicit domain artifacts (Table~\ref{tab:authoring_compact}), including executable rules and, when needed, a classifier.
\\
\textit{\textbf{LLM-based table extraction and document parsing.}}\looseness-1{} \textsc{InstrucTE}~\cite{bai2024schema}, LLM data-integration systems~\cite{steiner2026integration}, table models such as TaPas and TURL~\cite{herzig2020tapas,deng2022turl}, operate primarily over structured or recovered tables; surveys discuss the broader use of language models with tabular data~\cite{li2023tabular,narayan2022structured}. For raw PDFs, layout-aware parsers~\cite{auer2024docling,wang2024mineru,blecher2023nougat} help, but no single parser is reliable across scientific layouts, so \textsc{ArticleMiner} treats their outputs as evidence to be validated.
\\
\textit{\textbf{Semantic lifting and ontology-guided KG construction.}}\looseness-1{} Mapping languages (R2RML, RML), OBDA, and SHACL lift and validate RDF from structured sources, and KG construction needs stable identifiers, datatypes, and provenance~\cite{das2012r2rml,dimou2014rml,xiao2018obda,knublauch2017shacl,hogan2021kg,noy2019industrial,wilkinson2016fair}. \textsc{ArticleMiner} sits one stage earlier, recovering units, null values, identifiers, and context before a structured source exists and exposing structured records for downstream RDF mappings.


\section{Discussion and Threats to Validity}
\label{sec:discussion}
The results support three conclusions within the evaluated tasks. For RQ1, the component removals show that explicit domain guidance contributes to extraction quality, with task- and backbone-dependent gains. For RQ2, canonicalization and task-defined identity make independently recovered candidates comparable, while source locators and backend support retain evidence for inspection. The backend ablations show that adding extractors does not consistently improve accuracy. For RQ3, the raw-PDF/pre-parsed gaps and omission-heavy settings identify acquisition as an important remaining limitation, whereas over-extraction on the smaller benchmarks also requires better candidate selection. These findings favor task-specific acquisition and correction policies over a universal configuration.
\\
\emph{\textbf{Adaptation and maintenance.}} A new task requires vocabulary and mappings, derivation and validation code, prompts, a row schema, identifiers, and RDF bindings; geochemistry also requires a deposit classifier. The four modules demonstrate reuse of the orchestration, not zero-shot transfer or a measured reduction in authoring time. We did not record person-hours. Maintenance requires domain expertise to distinguish vocabulary gaps from extraction errors, assess source-specific conventions, and revise artifacts when the schema changes. Logged rejections help prioritize inspection but do not automatically diagnose every cause. Adaptation consequently involves both domain specification and assessment of representative publications. The authoring inventory makes the required artifacts explicit: a vocabulary entry addresses a naming gap, a derivation encodes a declared relationship, and a validator restricts admissible outputs. None can substitute for evidence absent from the recovered document. The four worked examples illustrate this division across tasks without implying that module size measures the difficulty of a new domain.
\\
\emph{\textbf{Implications for publication-to-KG systems.}} Downstream reuse requires units, censoring status, identity, and evidence locators alongside canonical labels. Identity deserves particular attention during adaptation: sample name alone can merge distinct analytical methods, while overly specific keys retain duplicates. Defining the target observation before expanding its vocabulary is therefore a practical design recommendation, although its effect on authoring effort remains unevaluated.
\\
\emph{\textbf{Construct validity.}} Strict tuple-$F_1$, sample-$F_1$, and the GeoChem composite score measure different properties. GeoChem's numerical and null tiers use matched rows; position-based matching can exploit numerical overlap, and missing gold metadata receives full credit. These choices can raise scores without improving end-to-end recall. The weights can change the ranking of nearby variants; per-tier scores are therefore necessary. No direct measure of downstream KG query quality is reported.
\\
\emph{\textbf{Internal validity.}} Full-system comparisons alter multiple stages, and GeoChem additionally differs in supplementary-file access and identifier handling. They do not isolate the effect of ontology grounding. Post-hoc configurations, differing ground-truth snapshots, backbone-specific token limits, and unavailable MinerU OCR weights further restrict causal interpretation. Component ablations should be interpreted within their documented run conditions.
\\
\emph{\textbf{External validity.}} The 163 publications represent four manually authored tasks and are dominated by DiSCoMaT. ChemTables and MLTables contain only 9 and 15 papers, respectively. GeoChem relies heavily on machine-readable supplements, and the system constructs per-publication records rather than resolving entities across publications. These conditions limit claims about unfamiliar scientific fields.
\\
\emph{\textbf{Conclusion validity.}} The paired summaries resample publications rather than individual tuples. Small samples yield imprecise estimates. The exact historical runs underlying the paired summaries are not fully identified by the release, so those summaries are descriptive and do not establish formal statistical significance or absence of an effect. Repeated-run variability was not characterized sufficiently to interpret very small differences. API drift and incomplete version pinning can hinder reproduction~\cite{sallou2024threats}. Possible training-data overlap remains unaudited; sharing a backbone does not eliminate contamination or its interaction with prompting~\cite{golchin2024timetravel}.

\section{Conclusion}
\label{sec:conclusion}
ArticleMiner separates publication-level extraction from explicit task-specific semantic artifacts. Across four authored tasks, the reported raw-publication results favor the complete pipeline over same-LLM few-shot prompting, while component and parser analyses show substantial task dependence. The framework's contribution is a reusable interface for acquiring, interpreting, and reconciling evidence under bounded domain constraints. General adaptation cost, controlled supplementary-aware baselines, and downstream graph quality remain open evaluation needs. Future work includes cross-paper entity resolution, adaptive evidence acquisition, and expert-reviewed module expansion from logged failures.

\paragraph*{Supplemental Material Statement:}\label{supp_mat}
Source code, ontology modules, prompts, output schemas, outputs for all eight backbones, ablation grids, and author-contributed \textsc{GeoChem} ground truth are available under CC-BY-4.0 at \url{https://github.com/Abrar2652/articleminer-iswc26}.
Few-shot outputs cover Sonnet~4.6 and open backbones; other closed-backbone results require regeneration using the released runner. Appendix~\ref{app:repro} provides the full manifest.

\section*{Declaration of use of Generative AI}
The authors used generative AI in two roles. As \emph{research instrumentation}, large language models (Claude, GPT-4o, Gemini, and open 7--8B models) are the object of study and part of the \textsc{ArticleMiner} pipeline, with identifiers and configurations reported in Sections~\ref{sec:methodology}--\ref{sec:experiments} and the appendices. As \emph{manuscript assistance}, a generative AI tool was used for grammar, spelling, rewording, and readability; the authors reviewed and edited all such text and take full responsibility for the content.

\section*{Acknowledgments}
This work was supported in part by the Defense Advanced Research Projects Agency (DARPA) under Contract No. 140D0426C0018. Any opinions, findings, and conclusions or recommendations expressed in this paper are those of the author(s) and do not necessarily reflect the views of DARPA or its Contracting Agent, the U.S. Department of the Interior, and no official endorsement should be inferred. The work of the first author was supported by the Viterbi Graduate School Fellowship from the USC Viterbi School of Engineering.

\bibliography{references}

@inproceedings{bai2024schema,
  title     = {Schema-Driven Information Extraction from Heterogeneous Tables},
  author    = {Bai, Fan and Kang, Junmo and Stanovsky, Gabriel and
               Freitag, Dayne and Dredze, Mark and Ritter, Alan},
  booktitle = {Findings of the Association for Computational Linguistics:
               EMNLP 2024},
  year      = {2024},
  pages     = {10252--10273},
  address   = {Miami, Florida, USA},
  publisher = {Association for Computational Linguistics},
  doi       = {10.18653/v1/2024.findings-emnlp.600},
  url       = {https://aclanthology.org/2024.findings-emnlp.600/},
}

@inproceedings{gupta2023discomat,
  title     = {{DiSCoMaT}: Distantly Supervised Composition Extraction from
               Tables in Materials Science Articles},
  author    = {Gupta, Tanishq and Zaki, Mohd and Khatsuriya, Devanshi and
               Hira, Kausik and Krishnan, N.~M.~Anoop and Mausam},
  booktitle = {Proceedings of the 61st Annual Meeting of the Association
               for Computational Linguistics (Volume 1: Long Papers)},
  year      = {2023},
  pages     = {13465--13483},
  address   = {Toronto, Canada},
  publisher = {Association for Computational Linguistics},
  doi       = {10.18653/v1/2023.acl-long.753},
  url       = {https://aclanthology.org/2023.acl-long.753/},
}

@inproceedings{knoblock2026minmod,
  title     = {Exploiting {LLMs} and Semantic Technologies to Build a
               Knowledge Graph of Historical Mining Data},
  author    = {Knoblock, Craig A. and Vu, Binh and Shbita, Basel and
               Chiang, Yao-Yi and Krishna, Pothula Punith and Lin, Xiao
               and Muric, Goran and Pyo, Jiyoon and Trejo-Sheu, Adriana
               and Ye, Meng},
  booktitle = {The Semantic Web -- ISWC 2025, Proceedings, Part II},
  series    = {Lecture Notes in Computer Science},
  volume    = {16141},
  year      = {2025},
  pages     = {451--471},
  publisher = {Springer},
  doi       = {10.1007/978-3-032-09530-5_26},
}

@inproceedings{vu2024sand,
  title     = {{SAND}: A Tool for Creating Semantic Descriptions of
               Tabular Sources},
  author    = {Vu, Binh and Knoblock, Craig A.},
  booktitle = {The Semantic Web: ESWC 2022 Satellite Events},
  series    = {Lecture Notes in Computer Science},
  volume    = {13384},
  year      = {2022},
  publisher = {Springer},
  doi       = {10.1007/978-3-031-11609-4_12},
}

@inproceedings{vu2025semantic,
  title     = {A Domain-Independent Approach for Semantic Table
               Interpretation},
  author    = {Vu, Binh and Knoblock, Craig A. and Lin, Fandel},
  booktitle = {The Semantic Web -- ISWC 2025, Proceedings, Part I},
  series    = {Lecture Notes in Computer Science},
  volume    = {16140},
  year      = {2025},
  pages     = {235--252},
  publisher = {Springer},
  doi       = {10.1007/978-3-032-09527-5_13},
}

@article{steiner2026integration,
  title     = {Automatic End-to-End Data Integration using Large Language
               Models},
  author    = {Steiner, Aaron and Bizer, Christian},
  journal   = {arXiv preprint arXiv:2603.10547},
  year      = {2026},
  url       = {https://arxiv.org/abs/2603.10547},
}

@inproceedings{jimenezruiz2020semtab,
  title     = {{SemTab} 2019: Resources to Benchmark Tabular Data to
               Knowledge Graph Matching Systems},
  author    = {Jim{\'e}nez-Ruiz, Ernesto and Hassanzadeh, Oktie and
               Efthymiou, Vasilis and Chen, Jiaoyan and Srinivas, Kavitha},
  booktitle = {The Semantic Web -- ESWC 2020, Proceedings},
  series    = {Lecture Notes in Computer Science},
  volume    = {12123},
  year      = {2020},
  pages     = {514--530},
  publisher = {Springer},
  doi       = {10.1007/978-3-030-49461-2_30},
  note      = {Canonical SemTab challenge resource paper; published in
               ESWC 2020 LNCS proceedings (not the Semantic Web Journal).},
}

@inproceedings{huynh2022dagobah,
  title     = {{DAGOBAH}: Table and Graph Contexts for Efficient Semantic
               Annotation of Tabular Data},
  author    = {Huynh, Viet-Phi and Liu, Jixiong and Chabot, Yoan and
               Deuz{\'e}, Fr{\'e}d{\'e}ric and Labb{\'e}, Thomas and
               Monnin, Pierre and Troncy, Rapha{\"e}l},
  booktitle = {Proceedings of the Semantic Web Challenge on Tabular Data
               to Knowledge Graph Matching (SemTab 2021) co-located with
               the 20th International Semantic Web Conference (ISWC 2021)},
  series    = {CEUR Workshop Proceedings},
  volume    = {3103},
  year      = {2021},
  pages     = {19--31},
  publisher = {CEUR-WS.org},
  url       = {https://ceur-ws.org/Vol-3103/paper2.pdf},
}

@inproceedings{ritze2015dbpedia,
  title     = {Matching {HTML} Tables to {DB}pedia},
  author    = {Ritze, Dominique and Lehmberg, Oliver and Bizer, Christian},
  booktitle = {Proceedings of the 5th International Conference on Web
               Intelligence, Mining and Semantics (WIMS 2015)},
  year      = {2015},
  pages     = {10:1--10:6},
  publisher = {ACM},
  doi       = {10.1145/2797115.2797118},
}

@inproceedings{herzig2020tapas,
  title     = {{TaPas}: Weakly Supervised Table Parsing via Pre-training},
  author    = {Herzig, Jonathan and Nowak, Pawe{\l} Krzysztof and
               M{\"u}ller, Thomas and Piccinno, Francesco and Eisenschlos,
               Julian Martin},
  booktitle = {Proceedings of the 58th Annual Meeting of the Association
               for Computational Linguistics (ACL 2020)},
  year      = {2020},
  pages     = {4320--4333},
  publisher = {Association for Computational Linguistics},
  doi       = {10.18653/v1/2020.acl-main.398},
  url       = {https://aclanthology.org/2020.acl-main.398/},
}

@article{deng2022turl,
  title     = {{TURL}: Table Understanding through Representation Learning},
  author    = {Deng, Xiang and Sun, Huan and Lees, Alyssa and Wu, You and
               Yu, Cong},
  journal   = {Proceedings of the VLDB Endowment},
  year      = {2020},
  volume    = {14},
  number    = {3},
  pages     = {307--319},
  publisher = {VLDB Endowment},
  doi       = {10.14778/3430915.3430921},
}

@inproceedings{li2023tabular,
  title     = {Table Pre-training: A Survey on Model Architectures,
               Pre-training Objectives, and Downstream Tasks},
  author    = {Dong, Haoyu and Cheng, Zhoujun and He, Xinyi and Zhou,
               Mengyu and Zhou, Anda and Zhou, Fan and Liu, Ao and
               Han, Shi and Zhang, Dongmei},
  booktitle = {Proceedings of the Thirty-First International Joint
               Conference on Artificial Intelligence (IJCAI 2022)},
  year      = {2022},
  pages     = {5426--5435},
  publisher = {IJCAI Organization},
  doi       = {10.24963/ijcai.2022/761},
  url       = {https://www.ijcai.org/proceedings/2022/761},
}

@article{narayan2022structured,
  title     = {Can Foundation Models Wrangle Your Data?},
  author    = {Narayan, Avanika and Chami, Ines and Orr, Laurel and
               Arora, Simran and R{\'e}, Christopher},
  journal   = {Proceedings of the VLDB Endowment},
  year      = {2022},
  volume    = {16},
  number    = {4},
  pages     = {738--746},
  publisher = {VLDB Endowment},
  doi       = {10.14778/3574245.3574258},
}

@article{hogan2021kg,
  title     = {Knowledge Graphs},
  author    = {Hogan, Aidan and Blomqvist, Eva and Cochez, Michael and
               d'Amato, Claudia and de Melo, Gerard and Gutierrez,
               Claudio and Kirrane, Sabrina and Labra Gayo, Jos{\'e} Emilio
               and Navigli, Roberto and Neumaier, Sebastian and
               Ngonga Ngomo, Axel-Cyrille and Polleres, Axel and
               Rashid, Sabbir M. and Rula, Anisa and Schmelzeisen, Lukas
               and Sequeda, Juan F. and Staab, Steffen and Zimmermann, Antoine},
  journal   = {ACM Computing Surveys},
  volume    = {54},
  number    = {4},
  pages     = {71:1--71:37},
  year      = {2021},
  publisher = {ACM},
  doi       = {10.1145/3447772},
}

@article{noy2019industrial,
  title     = {Industry-scale Knowledge Graphs: Lessons and Challenges},
  author    = {Noy, Natalya F. and Gao, Yuqing and Jain, Anshu and
               Narayanan, Anant and Patterson, Alan and Taylor, Jamie},
  journal   = {Communications of the ACM},
  volume    = {62},
  number    = {8},
  pages     = {36--43},
  year      = {2019},
  publisher = {ACM},
  doi       = {10.1145/3331166},
}

@article{wilkinson2016fair,
  title     = {The {FAIR} Guiding Principles for Scientific Data
               Management and Stewardship},
  author    = {Wilkinson, Mark D. and Dumontier, Michel and Aalbersberg,
               IJsbrand J.~J. and Appleton, Gabrielle and Axton, Myles
               and Baak, Arie and Blomberg, Niklas and Boiten, Jan-Willem
               and da Silva Santos, Luiz Bonino and Bourne, Philip E.
               and others},
  journal   = {Scientific Data},
  volume    = {3},
  pages     = {160018},
  year      = {2016},
  publisher = {Nature Publishing Group},
  doi       = {10.1038/sdata.2016.18},
}

@article{swain2016chemdataextractor,
  title     = {{ChemDataExtractor}: A Toolkit for Automated Extraction of
               Chemical Information from the Scientific Literature},
  author    = {Swain, Matthew C. and Cole, Jacqueline M.},
  journal   = {Journal of Chemical Information and Modeling},
  volume    = {56},
  number    = {10},
  pages     = {1894--1904},
  year      = {2016},
  publisher = {American Chemical Society},
  doi       = {10.1021/acs.jcim.6b00207},
}

@techreport{hofstra2021cmio,
  title     = {Deposit Classification Scheme for the Critical Minerals
               Mapping Initiative Global Geochemical Database},
  author    = {Hofstra, Albert H. and Lisitsin, Vladimir and Corriveau,
               Louise and Paradis, Suzanne and Peter, Jan M. and
               Lauzi{\`e}re, Kathleen and Lawley, Christopher J.~M. and
               Gadd, Michael and Pilote, Jean-Luc and Honsberger, Ian
               and Bastrakov, Evgeniy and Champion, David C. and
               Czarnota, Karol and Doublier, Michael P. and Huston,
               David L. and Raymond, Oliver and VanDerWielen, Simon
               and Emsbo, Poul and Granitto, Matthew and Kreiner,
               Douglas C.},
  institution = {U.S. Geological Survey},
  year      = {2021},
  type      = {USGS Open-File Report},
  number    = {2021--1049},
  doi       = {10.3133/ofr20211049},
  url       = {https://pubs.usgs.gov/of/2021/1049/ofr20211049.pdf},
}

@misc{auer2024docling,
  title         = {Docling Technical Report},
  author        = {Auer, Christoph and Lysak, Maksym and Nassar, Ahmed and
                   Dolfi, Michele and Livathinos, Nikolaos and Vagenas,
                   Panos and Ramis, Cesar Berrospi and Omenetti, Matteo
                   and Lindlbauer, Fabian and Dinkla, Kasper and Morin,
                   Lokesh and Staar, Peter W.~J.},
  year          = {2024},
  eprint        = {2408.09869},
  archivePrefix = {arXiv},
  primaryClass  = {cs.CL},
  url           = {https://arxiv.org/abs/2408.09869},
}

@misc{wang2024mineru,
  title         = {{MinerU}: An Open-Source Solution for Precise Document
                   Content Extraction},
  author        = {Wang, Bin and Xu, Chao and Zhao, Xiaomeng and Ouyang,
                   Linke and Wu, Fan and Zhao, Zhiyuan and Xu, Rui and
                   Liu, Kaiwen and Qu, Yuan and Shang, Fukai and Zhang,
                   Bo and Wei, Liqun and Sui, Zhihao and Zhou, Wei and
                   Xiang, Botian and Wei, Runyu and Li, Renqiu and Su,
                   Xinyue and Zhang, Ka{\"\i}kang and Yan, Guangsan and
                   Zhang, Zhen and Sun, Yuantao and Liu, Zichao and
                   Peng, He and Wu, Haodong and Liu, Haohui and Yan,
                   Conghui and He, Conghui},
  year          = {2024},
  eprint        = {2409.18839},
  archivePrefix = {arXiv},
  primaryClass  = {cs.CV},
  url           = {https://arxiv.org/abs/2409.18839},
}

@inproceedings{madaan2023selfrefine,
  title     = {{Self-Refine}: Iterative Refinement with Self-Feedback},
  author    = {Madaan, Aman and Tandon, Niket and Gupta, Prakhar and
               Hallinan, Skyler and Gao, Luyu and Wiegreffe, Sarah and
               Alon, Uri and Dziri, Nouha and Prabhumoye, Shrimai and
               Yang, Yiming and Gupta, Shashank and Majumder, Bodhisattwa
               Prasad and Hermann, Katherine and Welleck, Sean and
               Yazdanbakhsh, Amir and Clark, Peter},
  booktitle = {Advances in Neural Information Processing Systems 36
               (NeurIPS 2023)},
  year      = {2023},
  url       = {https://proceedings.neurips.cc/paper_files/paper/2023/hash/91edff07232fb1b55a505a9e9f6c0ff3-Abstract-Conference.html},
}

@inproceedings{golchin2024timetravel,
  title     = {Time Travel in {LLMs}: Tracing Data Contamination in Large
               Language Models},
  author    = {Golchin, Shahriar and Surdeanu, Mihai},
  booktitle = {The Twelfth International Conference on Learning
               Representations (ICLR 2024)},
  year      = {2024},
  url       = {https://openreview.net/forum?id=2Rwq6c3tvr},
}

@misc{blecher2023nougat,
  title         = {{Nougat}: Neural Optical Understanding for Academic
                   Documents},
  author        = {Blecher, Lukas and Cucurull, Guillem and Scialom, Thomas
                   and Stojnic, Robert},
  year          = {2023},
  eprint        = {2308.13418},
  archivePrefix = {arXiv},
  primaryClass  = {cs.LG},
  url           = {https://arxiv.org/abs/2308.13418},
}

@inproceedings{xiao2018obda,
  title     = {Ontology-Based Data Access: A Survey},
  author    = {Guohui Xiao and Diego Calvanese and Roman Kontchakov and Domenico Lembo and Antonella Poggi and Riccardo Rosati and Michael Zakharyaschev},
  booktitle = {Proceedings of the Twenty-Seventh International Joint Conference on
               Artificial Intelligence, {IJCAI-18}},
  publisher = {International Joint Conferences on Artificial Intelligence Organization},
  pages     = {5511--5519},
  year      = {2018},
  month     = {7},
  doi       = {10.24963/ijcai.2018/777},
  url       = {https://doi.org/10.24963/ijcai.2018/777},
}

@misc{das2012r2rml,
  title        = {{R2RML}: {RDB} to {RDF} Mapping Language},
  author       = {Das, Souripriya and Sundara, Seema and Cyganiak, Richard},
  year         = {2012},
  howpublished = {{W3C Recommendation}},
  url          = {https://www.w3.org/TR/r2rml/}
}

@inproceedings{dimou2014rml,
  title     = {{RML}: A Generic Language for Integrated {RDF} Mappings of Heterogeneous Data},
  author    = {Dimou, Anastasia and Vander Sande, Miel and Colpaert, Pieter and Verborgh, Ruben and Mannens, Erik and Van de Walle, Rik},
  booktitle = {Proceedings of the 7th Workshop on Linked Data on the Web},
  year      = {2014},
  url       = {https://ceur-ws.org/Vol-1184/ldow2014_paper_01.pdf}
}

@misc{knublauch2017shacl,
  title        = {{Shapes Constraint Language} ({SHACL})},
  author       = {Knublauch, Holger and Kontokostas, Dimitris},
  year         = {2017},
  howpublished = {{W3C Recommendation}},
  url          = {https://www.w3.org/TR/shacl/}
}

@inproceedings{sallou2024threats,
  author    = {June Sallou and Thomas Durieux and Annibale Panichella},
  title     = {Breaking the Silence: the Threats of Using {LLMs} in Software Engineering},
  booktitle = {Proceedings of the 2024 ACM/IEEE 46th International Conference on Software Engineering: New Ideas and Emerging Results (ICSE-NIER)},
  year      = {2024},
  pages     = {102--106},
  doi       = {10.1145/3639476.3639764}
}

\clearpage
\appendix

\section{The Three Integration Points of the Ontology Module}
\label{app:integration}

The pipeline consults $\Omega_d$ (Figure~\ref{fig:module}) at three points, linking evidence to the declared task representation: prompt-time grounding at extraction, canonicalization and validation at reconciliation, and IRI binding at serialization (Figure~\ref{fig:architecture}).

\emph{Prompt-time grounding} (integration point \textcircled{1}): $V_d$ enters the LLM's system prompt at extraction time. Admissible classes, canonical unit strings, and admissible null kinds are rendered into the prompt as an enumeration; the LLM is instructed to emit rows drawn only from those enumerations. The effect is to restrict the LLM's hypothesis space so that surface variations (e.g., \texttt{`micromolar'} vs \texttt{`\textmu M'}) map to canonical forms before $C_d$ ever runs.

\emph{Canonicalization and per-tuple validation} (integration point \textcircled{2}): $\Phi_d$, $C_d$, and $\Gamma_d$ run during reconciliation in the order specified by Eq.~\eqref{eq:rdef}. Every canonicalized candidate is evaluated against the executable predicates in $\Gamma_d$; those that fail are excluded from the trusted graph with a structured rejection reason (e.g., \texttt{unit-not-admissible}, \texttt{sum-not-100}, \texttt{summary-row}). Rejection reasons and evidence locators are logged so an auditor can trace why a record did not appear in $R(P,d)$.

\emph{IRI and \texttt{@context} declaration} (integration point \mbox{\textcircled{3}}): the RDF binding specifies how output fields are named in the graph. The released JSON-LD exporter declares a vocabulary and emits sample nodes with measurements and source metadata. This does not establish that every generated identifier resolves to an externally maintained ontology term, or that a native equivalent Turtle export has been evaluated (Appendix~\ref{app:jsonld}).

The three points are functionally distinct: (\textcircled{1}) guides what the LLM proposes; (\textcircled{2}) canonicalizes, derives, and rejects records that fail structural constraints; (\textcircled{3}) fixes the RDF identity of what remains. These integration points have different intended roles: prompt grounding guides proposed fields; record checks enforce declared constraints; and RDF bindings define the graph vocabulary. These roles do not constitute separate empirical ablations of each integration point.

\section{Cross-Domain Worked Example Analogs}
\label{app:cross-domain-worked}

The geochemistry worked example in \S\ref{sec:method-walkthrough} traces one row (\texttt{BR-12} with \texttt{Au<0.5} and method \texttt{ICP-MS}) through Parse, Extract, Reconcile, and Serialize. The following schematic examples expand the cross-domain analogs in the main text; they are not additional evaluated outputs or verbatim serializer dumps.

\paragraph{Materials science (\textsc{DiSCoMaT}).}
A glass-composition row reports \texttt{SiO}$_2$: 55, \texttt{Na}$_2$\texttt{O}: 25, \texttt{CaO}: 20 in a table whose caption declares the unit as \texttt{mol\%}. \textbf{Parse} recovers the row and its header via docling. \textbf{Extract} emits the provisional record $\tilde r = (\text{material}, [(\text{SiO}_2, 55), (\text{Na}_2\text{O}, 25), (\text{CaO}, 20)], \text{unit=mol\%})$. \textbf{Reconcile}: $\Phi_d$ is a no-op here (no taxonomic derivation on the constituents), $C_d$ passes the oxide names through (already canonical), and $\Gamma_d$ accepts because (i) all constituents are in the oxide vocabulary, (ii) the unit is admissible, and (iii) $\sum_i p_i = 100$ within tolerance. \textbf{Serialize} must retain each constituent, percentage, and declared unit. A row whose caption instead declares wt\% represents a different composition and cannot be substituted for the mol\% row in a downstream query. This example specifies the information to preserve without assuming an external oxide or unit IRI mapping.

\paragraph{Drug-discovery chemistry (\textsc{ChemTables}).}
A bioactivity table reports \texttt{IC50 >10} for compound \texttt{7a} against \texttt{HER2}, with a header declaring $\mu$M. The extracted record must preserve the treatment, target, assay, unit, and inequality. Here $>10$ is a lower bound on the concentration, not a measured value of 10 or an upper bound. A downstream query for compounds with IC$_{50}\leq5\,\mu$M must exclude this observation. The evaluated prompt retains the literal inequality; an RDF mapping may encode its comparator and threshold explicitly.

\paragraph{Machine learning (\textsc{MLTables}).}
An arXiv table reports \texttt{Ours: 87.3} under \texttt{Accuracy (\%)}, with \texttt{CIFAR-10} identified in the caption. Extraction links the value to its model, dataset, metric, and unit using table and prose evidence. Resolving \texttt{Ours} requires source context at extraction time; it is not a lookup that a fixed canonicalizer can perform without that evidence. Subsequent normalization and validation enforce the declared task schema.

The examples distinguish contextual interpretation by the extraction model from deterministic operations specified by the module. Across tasks, semantic fields and schemas change while the high-level orchestration is reused.

\section{Output Schemas}
\label{app:schema}

The \textsc{ArticleMiner} \textsc{GeoChem} ground-truth schema contains 209 columns: 63 metadata/provenance fields and 146 element columns (73 elements $\times$ \{value, detection-limit/null information\}). The pipeline's output Excel allows extra optional fields (uncertainty, method, alternative units), but evaluation pulls only the named-field projection (\S\ref{sec:evaluation}), so additional pipeline-emitted columns are ignored rather than penalized, and pipeline and baseline are compared on the same 209-column projection. The other three benchmarks each define a tuple shape: ChemTables uses $\langle\textit{cell\_index, value, type, target, treatment, unit}\rangle$ with $\textit{type}\in\{\text{IC}_{50},\text{EC}_{50},\text{GI}_{50},\text{CC}_{50},\text{MIC}\}$; DiSCoMaT uses $\langle\textit{sample\_id, component, percentage, unit}\rangle$ with the unit in \{mol\%, wt\%\}; MLTables uses an \emph{entry-type}-tagged tuple in $\{\text{Result}, \text{Data Stat.}, \text{Hyper-parameter}, \text{Other}\}$ with type-dependent attribute fields (e.g.\ Result carries \textit{model, dataset, metric, task}).

\subsection{GeoChem scoring details}
\label{app:metric-details}
For $T_1$, string comparison returns the maximum of token Jaccard similarity and sequence similarity after normalization. Missing gold values score 1; a missing prediction for a present gold value scores 0. For a nonzero numerical gold value $g$, let $e=|p-g|/|g|$. The released $T_2$ score is 1 for $e\leq0.05$, 0 for $e\geq1$, and $1-(e-0.05)/0.95$ otherwise. A zero gold value receives full credit only within absolute error 0.001. When the gold is the below-detection sentinel, a matching sentinel scores 1, a missing prediction 0.5, and another value 0.3. A predicted sentinel against an ordinary gold value scores 0. $T_2$ averages scored fields within a matched row and then averages matched rows; $T_4$ similarly averages null agreement over gold-null fields.

Identifier matching tests available sample-name columns and accepts exact or prefix matches. If fewer than 30\% of gold rows match by name, a fallback compares row order under offsets and numerical overlap. Rows are compatible when at least 30\% of shared positive elemental values agree within 10\%; for much larger prediction sets, a greedy value-fingerprint match is used. The fallback therefore uses measurement values to establish alignment, which must be considered when interpreting numerical accuracy. $T_3$ is the resulting sample-$F_1$, while row counts are also reported diagnostically. These rules describe the released evaluator; they do not make its score equivalent to strict tuple matching.

\section{Per-Domain External-Benchmark Detail}
\label{app:external}

\paragraph{DiSCoMaT.}
On raw PDFs, Opus~4.6 reaches 81.0 tuple-$F_1$ and Sonnet~4.6 reaches 79.3, compared with same-backbone few-shot scores of 59.5 and 64.5. Sonnet's 98.2\% precision and 66.4\% recall indicate that omissions dominate the remaining error. The reported per-paper analysis includes 32 of 111 papers with no predictions; these failures remain in the main denominator. Published GNN results are external context and do not share the paired analysis in Table~\ref{tab:significance}.

\paragraph{MLTables.}
Opus~4.6 reaches 55.9 tuple-$F_1$ against its few-shot baseline's 41.6. The Sonnet paired means in Table~\ref{tab:significance}(b) are a separate comparison: 52.2 versus 37.8, with a reported paired difference of 14.4. Open-model scores range from 28.6 to 51.1. These observations are limited by the 15-paper sample.

\paragraph{ChemTables.}
The full Sonnet configuration reaches 32.1 tuple-$F_1$ against 15.8 for few-shot prompting; disabling paper intelligence raises the system score to 44.6. The latter is a post-hoc configuration. The 9-paper paired summary has wide intervals. The configuration underlying the historical paired summary is not fully documented; that summary should not be treated as a confirmed paired test of the full pipeline.

\paragraph{GeoChem.}
Closed-backbone composite scores range from 75.5 to 76.3 on 28 entries. The separate paired analysis uses 25 common entries and reports a mean sample-$F_1$ of 63.2 for Sonnet. These values describe different estimands and denominators. Supplementary-file access and sample-identifier handling contribute to the gap from PDF-only few-shot prompting.

\section{Ontology Module Entries}\label{app:module-entries}
\label{app:ontology}

This appendix gives the full per-task semantic content behind the compact summary in Table~\ref{tab:authoring_compact}. Representative entries from the four domain-specific \textsc{ArticleMiner} ontology modules follow; full modules are released with the source code.

\subsection{Geochemistry (220 entries)}
The geochemistry module combines deposit keyword mappings, mineral labels, analytical-method aliases, and value conventions. Approximately 90 keyword entries map deposit descriptions into the adopted hierarchy; these mappings are distinct from the complete 189-type classifier vocabulary. Around 80 mineral entries associate labels with coarse classes and formulae, for example sphalerite with sulfide and ZnS, or magnetite with oxide and Fe$_3$O$_4$. More than 50 method variants map to approximately 12 canonical methods, including LA-ICPMS, EPMA, SIMS, ICP-OES, and XRF. Value rules distinguish detection-limit encodings from missing values and preserve explicit unit classes. These counts describe different artifact categories and should not be interpreted as the number of axioms in a formal ontology.

\subsection{Drug discovery: ChemTables (80 entries)}
The bioassay vocabulary includes five activity types \{IC$_{50}$, EC$_{50}$, GI$_{50}$, MIC, CC$_{50}$\} together with their canonical units and assay interpretations. The target taxonomy lists representative protein kinases (CHK1, CHK2, EGFR, $\ldots$), receptors, and microbial targets seen across the 9 PMC papers, with synonym maps where applicable. The module includes assay-specific value ranges and compound-target linkage constraints; these do not establish monotonicity of an unobserved dose-response curve.

\subsection{Materials science: DiSCoMaT (167 entries)}
The material-classification taxonomy distinguishes glass, ceramic, alloy, polymer, plus a long tail of sub-classes from the SciGlass database. The oxide-component taxonomy lists the standard compositional building blocks (SiO$_2$, Na$_2$O, B$_2$O$_3$, PbO, CaO, $\ldots$) with their canonical chemical formulae. Unit standardization distinguishes mol\% vs.\ wt\% (both valid; the distinction must be preserved per table). Constraints encode composition-sum tolerance ($\sum\!\approx\!100\%$) and component plausibility ranges per material class.

\subsection{Machine learning: MLTables (60 entries)}
The entry-type taxonomy enumerates the four MLTables annotation classes \{Result, Data Statistic, Hyper-parameter/Architecture, Other\}. The metric taxonomy lists common ML metrics (accuracy, F$_1$, BLEU, perplexity, AUC, mAP, $\ldots$) with their value ranges and interpretation (higher-better vs.\ lower-better). Attribute linking defines the model$\to$dataset$\to$metric$\to$value chain that identifies a Result and the parameter$\to$model association that identifies a Hyper-parameter. Constraints encode metric plausibility (accuracy $\in[0,100]$, loss $\geq 0$) and model-dataset consistency within a paper.

\section{Prompt Templates}
\label{app:prompts}

This appendix documents the prompt templates used in each LLM-based stage of \textsc{ArticleMiner}; full templates are released with the source code. Listings~1--4 reproduce the extraction system prompt of each task module \emph{verbatim} from the released code, so the per-domain differences are directly comparable: the prompts share a high-level structure (role, target output, domain rules, and prohibitions on fabrication), while the vocabulary, tuple shape, unit conventions, and null-value semantics come from $\Omega_d$. Note that the dash convention is deliberately opposite across domains: in \textsc{DiSCoMaT} composition tables, a \texttt{-} cell means the component is absent ($0\%$), whereas the GeoChem task default treats it as below detection; source legends may require a different interpretation (\S\ref{sec:method-geochem-knowledge}, Part~3). The same surface symbol carries different semantics per domain, which is why null conventions live in the task module rather than in shared code.

\begin{tcolorbox}[enhanced, breakable, colback=amBlueL!40, colframe=amBlue,
  boxrule=0.4pt, arc=2pt, left=6pt, right=6pt, top=4pt, bottom=4pt,
  title={\scriptsize\textbf{Listing 1: Exact \textsc{DiSCoMaT} extraction system prompt}},
  fonttitle=\color{white}, coltitle=white, colbacktitle=amBlue]
\label{lst:discomat-prompt}
{\footnotesize\ttfamily\raggedright
You are an expert materials scientist specializing in glass and ceramic compositions. Your task is to extract material composition data from tables in materials science papers.\\[3pt]
\#\# Extraction Target\\
For each composition table, extract tuples of the form:\\
\{\,"sample\_id": "<identifier, from row header or sample number>",\\
\phantom{\{}\,"component": "<chemical formula, e.g., SiO2, Na2O, PbO>",\\
\phantom{\{}\,"value": <numerical percentage value>,\\
\phantom{\{}\,"unit": "<mol or wt>"\,\}\\[3pt]
\#\# Unit Inference Rules\\
1. Check the table caption for "mol\%", "mole\%", "wt\%", or "weight\%".\\
2. Check column headers for unit annotations like "SiO2 (mol\%)".\\
3. If the caption says "mol\%" then ALL values in the table are mol\%.\\
4. If no unit is found, check if the values sum to ${\sim}100\%$.\\[3pt]
\#\# Critical Rules\\
1. ONLY extract composition data; skip physical properties (density, $T_g$, hardness).\\
2. Component names must be valid chemical formulas (SiO2, not "silica").\\
3. A cell of "-" or "--" means 0\% (component absent); do NOT extract it.\\
4. A blank cell means "not reported"; do NOT extract it.
}
\end{tcolorbox}

\begin{tcolorbox}[enhanced, breakable, colback=amBlueL!40, colframe=amBlue,
  boxrule=0.4pt, arc=2pt, left=6pt, right=6pt, top=4pt, bottom=4pt,
  title={\scriptsize\textbf{Listing 2: Exact \textsc{GeoChem} raw-PDF table extraction system prompt}},
  fonttitle=\color{white}, coltitle=white, colbacktitle=amBlue]
\label{lst:geochem-prompt}
{\footnotesize\ttfamily\raggedright
You are an expert geochemistry database curator. You are given the text of a research paper (including any tables embedded in the PDF). There is NO supplementary spreadsheet available. Your task is to extract individual sample analytical data (element concentrations) directly from tables in the PDF text.\\[3pt]
Extract information EXACTLY as reported --- do not paraphrase, estimate, or fabricate values.\\[3pt]
\{knowledge\_base\}\\[3pt]
\textrm{\textit{The \texttt{\{knowledge\_base\}} placeholder is where $V_d$ is rendered into the prompt at run time (integration point \textcircled{1}): the admissible element list, unit classes, mineral and method vocabularies, and the null-kind conventions (\texttt{BDL}, \texttt{NM}, measured zero). The companion user template requests a JSON object with \texttt{samples} and \texttt{extraction\_notes}, one object per analytical spot.}}
}
\end{tcolorbox}

\begin{tcolorbox}[enhanced, breakable, colback=amBlueL!40, colframe=amBlue,
  boxrule=0.4pt, arc=2pt, left=6pt, right=6pt, top=4pt, bottom=4pt,
  title={\scriptsize\textbf{Listing 3: Exact \textsc{ChemTables} extraction system prompt}},
  fonttitle=\color{white}, coltitle=white, colbacktitle=amBlue]
\label{lst:chemtables-prompt}
{\footnotesize\ttfamily\raggedright
You are an expert medicinal chemist specializing in drug discovery data extraction. Your task is to extract bioactivity measurements from tables in medicinal chemistry papers.\\[3pt]
\#\# Extraction Target\\
For each numerical cell in the table that represents a bioactivity measurement, produce a JSON object:\\
\{\,"cell\_index": "CA(row,col)",\\
\phantom{\{}\,"value": "<exact numerical value from the cell>",\\
\phantom{\{}\,"type": "<one of: IC50, EC50, GI50, MIC>",\\
\phantom{\{}\,"target": "<protein target, cell line, or organism being assayed>",\\
\phantom{\{}\,"treatment": "<compound identifier (number, name, or code) being tested>",\\
\phantom{\{}\,"unit": "<measurement unit, e.g., $\mu$M, nM, $\mu$g/mL>"\,\}\\[3pt]
\#\# Assay Type Definitions\\
- IC50: Half-maximal inhibitory concentration (enzyme/protein inhibition AND cell viability/proliferation assays)\\
- EC50: Half-maximal effective concentration (cell-based functional assays)\\
- GI50: Concentration causing 50\% growth inhibition (ONLY when paper explicitly uses the term GI50)\\
- MIC: Minimum inhibitory concentration (antimicrobial assays)\\[3pt]
IMPORTANT: If the table caption or column header explicitly states "IC50", classify as IC50 even if the assay measures cell proliferation. The paper's own terminology takes precedence over domain conventions. Only use GI50 if the paper explicitly labels values as GI50.\\[3pt]
\#\# Critical Rules\\
1. Extract ALL cells containing numerical bioactivity/potency values. This includes IC50, EC50, GI50, MIC, and related measurements like \% inhibition (which should be classified as MIC if units are $\mu$g/mL or $\mu$M) or cytotoxicity values.\\
2. The "treatment" is the compound being tested --- usually identified by a number in the first column or a compound name.\\
3. The "target" is what the compound is tested against --- a protein (CHK1, EGFR), cell line (A549, HeLa), or organism (S. aureus, E. coli). If the target is not clear, use "xx".\\
4. Infer type, target, and unit from column headers. Column headers often follow patterns like "CHK1 IC50 ($\mu$M)" or "MIC ($\mu$g/mL) S. aureus".\\
5. If a column header says "\% inhibition" or "inhibition (\%)" with units like $\mu$g/mL, classify as MIC.\\
6. If a cell contains a range like "1.0 (0.86, 1.2)", extract the primary value "1.0" AND include the full string with $\pm$ if present (e.g., "265 $\pm$ 17" should have value "265 $\pm$ 17").\\
7. If a cell contains ">100" or ">50", this means the compound is inactive at the tested range. Still extract it with value ">100" or ">50".\\
8. Do NOT hallucinate values. Only extract what is explicitly present in the table.\\
9. Skip truly non-bioactivity data like selectivity ratios, ligand efficiency, or physical properties (LogP, MW).\\
10. Output one JSON object per line. No markdown, no explanation.
}
\end{tcolorbox}

\begin{tcolorbox}[enhanced, breakable, colback=amBlueL!40, colframe=amBlue,
  boxrule=0.4pt, arc=2pt, left=6pt, right=6pt, top=4pt, bottom=4pt,
  title={\scriptsize\textbf{Listing 4: Exact \textsc{MLTables} extraction system prompt}},
  fonttitle=\color{white}, coltitle=white, colbacktitle=amBlue]
\label{lst:mltables-prompt}
{\footnotesize\ttfamily\raggedright
You are an expert ML researcher. Given a table extracted from a machine learning paper, extract ALL quantitative entries (results, dataset statistics, hyperparameters).\\[3pt]
For each entry, output a JSON object on one line:\\
\{"value": "<exact number>", "type": "<Result$|$Data Stat.$|$Hyper-parameter/Architecture$|$Other>", \ldots attributes\ldots\}\\[3pt]
For type "Result": include "model", "dataset", "metric", "task" if identifiable.\\
For type "Data Stat.": include "dataset", "attribute name" (e.g., "number of samples", "number of classes").\\
For type "Hyper-parameter/Architecture": include "parameter/architecture name", "model".\\[3pt]
RULES:\\
1. Extract the EXACT numerical value from the table cell.\\
2. Classify each value into one of the four types based on context.\\
3. For Results: the model is usually in the row, the metric in the column header, the dataset from caption or headers.\\
4. Skip non-quantitative cells (model names, dataset names used as labels).\\
5. Do NOT hallucinate. Only extract what is explicitly in the table.\\
6. Output ONLY JSON lines, no explanation.
}
\end{tcolorbox}

The verbatim ChemTables prompt enumerates four assay labels although the module also includes CC$_{50}$. Its rule mapping some inhibition measurements to MIC is a task heuristic, not a biochemical equivalence. Likewise, a composition summing to 100 does not distinguish mol\% from wt\%. These prompt choices are retained to document the evaluated instrumentation and can introduce systematic interpretation errors.

\paragraph{Paper-intelligence Blueprint skim (planning stage).}
The blueprint prompt asks the model to identify analytical methodology without inventing values. It receives PDF text and the task vocabulary, and returns expected sample count, measured elements, minerals, analytical methods, instrument, laboratory, standards, and operating conditions. The expected count guides acquisition retries; the other fields provide extraction context.

\paragraph{Paper-level metadata LLM (planning stage).}
System role: extract paper-level metadata, validate each category against the provided taxonomy, and reject categories absent from it; never invent units or method names. User input: PDF text, the Blueprint object, and the classification taxonomy (e.g.\ the 189-class CMMI deposit-type list for geochemistry). Output: JSON conforming to the paper-level metadata schema, with per-category confidence.

\paragraph{Agentic diagnostic LLM (structural-hint retry).}
System role: diagnose why parsing produced fewer rows than expected and return structured hints for a retry, enumerating the failure modes to look for (wrong header row, transposed layout, unrecognized sample-ID column, wrong unit assumption, multi-row headers, non-geochemical content). It requests only parsing hints, not values --- the iterative refine--feedback pattern of~\cite{madaan2023selfrefine} restricted to structural correction. User input: failed-extraction rows, expected sample count, raw table text per backend. Output: JSON hints deserialized into \texttt{\{header\_row, is\_transposed, sample\_id\_col, element\_label\_col, notes\_from\_llm\}}, with \texttt{null} for fields whose auto-detection was correct; separate supplement-file and raw-PDF variants exist, bounded at two diagnosis attempts per failing file.

Token budgets are set per stage rather than globally (most extraction and diagnosis calls use $8192$; short classification and filtering calls use less); temperature is $0.0$ throughout for reproducibility.

\paragraph{Open-LLM cap.} For 7--8B open backbones (Qwen-2.5, Mistral-v0.2, Llama-3.1) on \textsc{GeoChem}, \\\texttt{max\_new\_tokens} is capped at 512 across blueprint, metadata, table-extraction, and deposit-classifier stages. The cap suppresses hallucination spirals observed under longer budgets (e.g., open-ended isotope-chain enumeration on geochemistry tables) and short-circuits failed JSON-schema completions; the deposit classifier still emits malformed JSON on the 189-class CMMI taxonomy under all three open backbones, which the pipeline catches and writes \texttt{deposit\_type=null} (\S\ref{sec:results-external}, ``Open-LLM geochemistry'').

\paragraph{Few-shot baseline prompt.}
The PDF few-shot baseline uses one call at temperature zero with three fixed exemplars. Exemplars come from the external benchmarks' training splits or a separate GeoChem exemplar pool. The user input contains PDF text, or released table text for the pre-parsed setting, without a blueprint, ontology context, vision, or retries. The reported token budgets are 4096 input and 8192 output tokens. The GeoChem baseline emits a compact per-row schema rather than the complete 209-column projection and does not ingest supplementary spreadsheets. Its comparison with ArticleMiner therefore includes input-access and representation differences. The ChemTables exemplar-count analysis motivates three exemplars for that task; it does not establish context sufficiency elsewhere.

\section{Sonnet 4.6 Sensitivity on PDF-only Domains}
\label{app:sonnet_sensitivity}

The main-text ablation table (Table~\ref{tab:ablation}, all panels) fixes the LLM backbone to Haiku 4.5 for cross-domain uniformity. On the small PDF-only benchmarks \textsc{ChemTables} ($n{=}9$) and \textsc{MLTables} ($n{=}15$), the magnitudes of component effects differ between Haiku and Sonnet. Re-running the same ablations on Sonnet 4.6 shows that component magnitudes and, for paper intelligence, the direction of the effect change with the backbone, so the two grids should be read as task- and backbone-specific diagnostics rather than as a single universal ranking.

\begin{table}[ht]
\centering\scriptsize
\caption{\textbf{Sonnet 4.6 component ablation} on \textsc{ChemTables}/\textsc{MLTables}, companion to Table~\ref{tab:ablation}. Vision is the largest single contributor on PDF-only domains ($-17.5$ / $-47.4$ F$_1$ when removed); removing paper-intelligence raises \textsc{ChemTables} F$_1$ by $+12.5$ via stricter precision. Bold = largest drop / largest gain.}
\label{tab:ablation_sonnet}
\begin{tabular}{lccc c ccc}
\toprule
 & \multicolumn{3}{c}{\textbf{ChemTables} ($n{=}9$)} & & \multicolumn{3}{c}{\textbf{MLTables} ($n{=}15$)} \\
\cmidrule(lr){2-4} \cmidrule(lr){6-8}
\textbf{Variant} & P & R & F$_1$ & & P & R & F$_1$ \\
\midrule
\textbf{Full pipeline} & 30.4 & 34.0 & 32.1 &  & 52.6 & 51.7 & 52.2 \\
w/o ontology           & 28.8 & 32.3 & 30.4 &  & 51.9 & 51.6 & 51.7 \\
w/o self-correction    & 28.6 & 32.3 & 30.3 &  & 51.8 & 51.6 & 51.7 \\
w/o validation         & 28.4 & 32.3 & 30.2 &  & 51.7 & 50.8 & 51.2 \\
w/o intelligence       & \textbf{64.9} & 34.0 & \textbf{44.6}$^{\uparrow}$ &  & 52.5 & 52.1 & 52.3 \\
w/o vision             & 17.1 & 12.8 & \textbf{14.6} &  & 28.3 &  2.6 & \phantom{0}\textbf{4.8} \\
\bottomrule
\end{tabular}
\\
{\footnotesize $^{\uparrow}$ removing the paper-intelligence stage raises \textsc{ChemTables} F$_1$ by $+12.5$, with precision increasing from 30.4 to 64.9 while recall remains 34.0; this identifies an interaction but does not establish its mechanism. Cross-model reading: effects are task- and backbone-dependent (e.g., removing paper intelligence lowers \textsc{ChemTables} F$_1$ under Haiku but raises it under Sonnet), so neither grid gives universal component effects; vision produces the largest losses under Sonnet, whereas validation removal produces larger losses under Haiku, and \textsc{GeoChem}'s supplementary-data path makes LLM-side components far less influential on that task.}
\end{table}

\section{Extended Results}
This appendix provides supporting model, prompt-count, and protocol analyses. Table~\ref{tab:geochem_main} is a historical sequence of configurations, not a controlled one-factor ablation: changes in $T_4$ cannot be assigned solely to the deposit classifier. Table~\ref{tab:cross_llm_agreement} describes overlap among model outputs rather than their correctness.

\begin{table}[!ht]
\centering
\footnotesize
\caption{Post-hoc protocol-component analysis on \textsc{GeoChem} (All-28, Sonnet~4.6): four-tier scores as protocol components are added. Exploratory diagnostic, not independent confirmation. \textbf{Bold} marks the best per column.}
\label{tab:geochem_main}
\setlength{\tabcolsep}{8pt}
\begin{tabular}{l ccccc}
\toprule
\textbf{Configuration} & $T_1$ & $T_2$ & $T_3$ & $T_4$ & \textbf{Overall $S$} \\
\midrule
Base configuration                     & 69.9 & 70.7 & 67.8 & 88.4 & 72.7 \\
\ + USGS protocol                      & 71.5 & 70.9 & \textbf{70.8} & 88.4 & 73.7 \\
\ + LLM deposit classifier (3-backend) & 71.2 & 74.0 & 68.5 & \textbf{97.6} & 75.9 \\
\ + full 5-backend (\textsc{ArticleMiner}) & \textbf{71.6} & \textbf{76.5} & 68.5 & 90.6 & \textbf{76.0} \\
\bottomrule
\end{tabular}
\end{table}

\begin{table}[!ht]
\centering
\scriptsize
\caption{Pairwise tuple Jaccard (\%) between backbones and their three-way intersection; higher = more agreement.}
\label{tab:cross_llm_agreement}
\setlength{\tabcolsep}{10pt}
\begin{tabular}{l cccc}
\toprule
\textbf{Backbone pair} & \textbf{\textsc{ChemTables}} & \textbf{\textsc{MLTables}} & \textbf{\textsc{DiSCoMaT}} & \textbf{\textsc{GeoChem}} \\
\midrule
Sonnet~4.6 / Opus~4.6  & 99 & 85 & 79 & 90 \\
Sonnet~4.6 / GPT-4o  & 72 & 57 & 74 & 79 \\
Opus~4.6 / GPT-4o   & 72 & 60 & 64 & 82 \\
Three-way    & 58 & 53 & 52 & 75 \\
\bottomrule
\end{tabular}
\end{table}

\begin{figure}[!ht]
\centering
\includegraphics[width=0.85\linewidth]{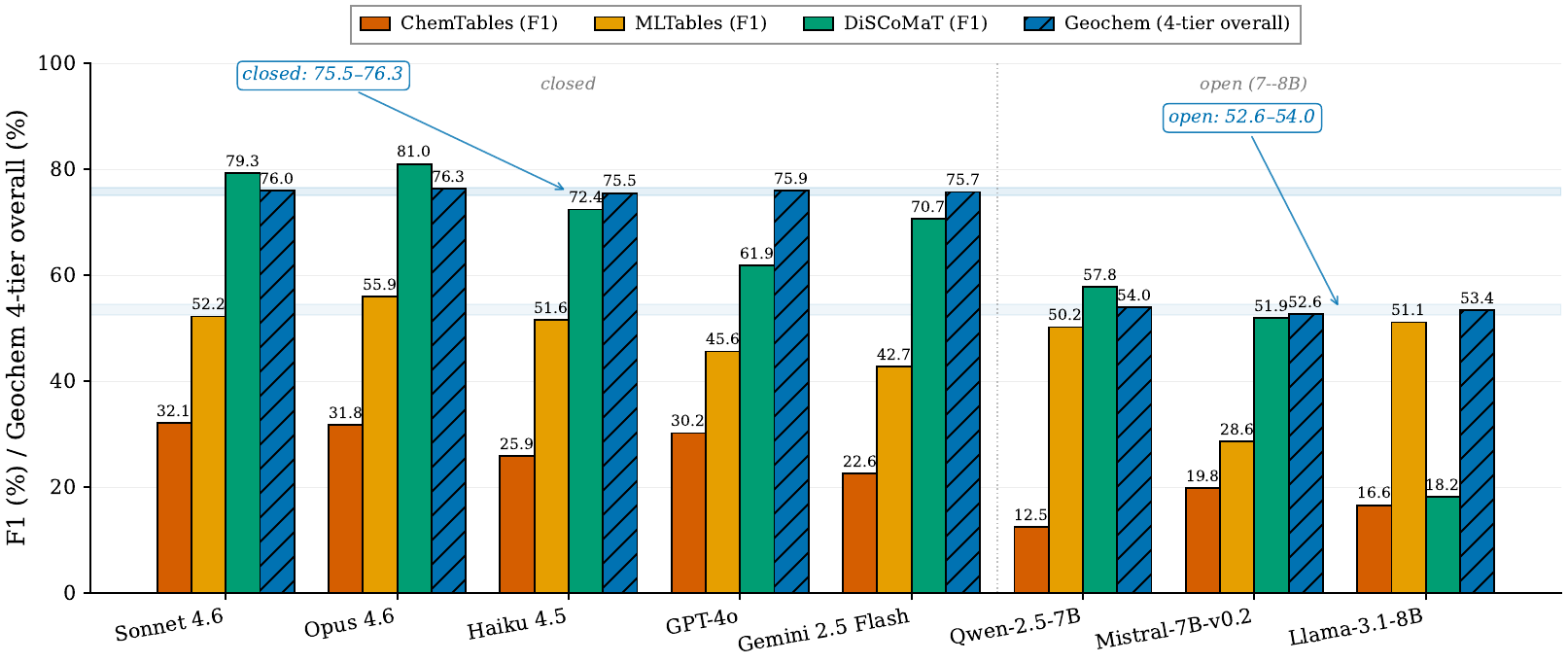}
\caption{Per-backbone score across the four tasks (corresponding to Table~\ref{tab:main_closed}): strict tuple-$F_1$ (\%) for \textsc{ChemTables}/\textsc{MLTables}/\textsc{DiSCoMaT} and four-tier Overall $S$ (\%) for \textsc{GeoChem}. The two metrics are distinct constructs and are not directly comparable across tasks.}
\label{fig:multi_llm}
\end{figure}

Figure~\ref{fig:multi_llm} shows a 0.8-point spread in GeoChem's composite score across closed backbones. This stability is consistent with the shared deterministic spreadsheet path, but does not imply that every metadata field or individual prediction is invariant to the backbone.

\label{app:results}

\subsection{Full cross-model grid}
Tables~\ref{tab:consolidated_all}--\ref{tab:grid_geochem} reports the complete grid of precision, recall, and \mbox{$F_1$} for all eight backbones under both input modalities; the main text retains only the matched closed-model comparison (Table~\ref{tab:main_closed}) and its uncertainty (Table~\ref{tab:significance}).
\begin{table}[!ht]
\centering\small
\caption{ChemTables P/R/$F_1$ (\%). $n=9$ publications; raw-PDF and pre-parsed inputs. ArticleMiner is abbreviated AM.}
\label{tab:consolidated_all}
\setlength{\tabcolsep}{6pt}
\begin{tabular}{llcccccc}
\toprule
& & \multicolumn{3}{c}{Raw PDF} & \multicolumn{3}{c}{Pre-parsed} \\
\cmidrule(lr){3-5}\cmidrule(lr){6-8}
Backbone & System & P & R & $F_1$ & P & R & $F_1$ \\
\midrule
Sonnet 4.6 & AM & 30.4 & 34.0 & \textbf{32.1} & 57.8 & 60.2 & 59.0 \\
 & Few-shot & 41.7 & 9.7 & 15.8 & 48.3 & 39.2 & 43.2 \\
Opus 4.6 & AM & 30.6 & 33.1 & 31.8 & 59.9 & 61.5 & 60.7 \\
 & Few-shot & 43.7 & 9.7 & 15.9 & 50.9 & 50.9 & 50.9 \\
Haiku 4.5 & AM & 19.3 & 39.4 & 25.9 & 55.0 & 57.1 & 56.1 \\
 & Few-shot & 42.1 & 9.7 & 15.8 & 48.4 & 45.9 & 47.1 \\
GPT-4o & AM & 28.2 & 32.5 & 30.2 & 55.0 & 52.4 & 53.7 \\
 & Few-shot & 62.0 & 9.5 & 16.5 & 45.4 & 43.1 & 44.2 \\
Gemini 2.5 Flash & AM & 16.5 & 35.9 & 22.6 & 61.3 & 61.3 & 61.3 \\
 & Few-shot & 24.0 & 10.2 & 14.3 & 47.6 & 44.8 & 46.2 \\
Qwen-2.5-7B (open) & AM & 11.9 & 13.2 & 12.5 & 79.0 & 53.9 & \textbf{64.1} \\
 & Few-shot & 37.7 & 5.0 & 8.8 & 44.7 & 28.6 & 34.9 \\
Mistral-7B (open) & AM & 22.3 & 17.7 & 19.8 & 16.9 & 19.0 & 17.9 \\
 & Few-shot & 6.7 & 6.5 & 6.6 & 36.0 & 20.8 & 26.3 \\
Llama-3.1-8B (open) & AM & 18.6 & 14.9 & 16.6 & 49.9 & 54.3 & 52.0 \\
 & Few-shot & 15.4 & 8.7 & 11.1 & 45.3 & 30.5 & 36.5 \\
\bottomrule
\end{tabular}
\end{table}

\begin{table}[!ht]
\centering\small
\caption{MLTables P/R/$F_1$ (\%). $n=15$ publications; raw-PDF and pre-parsed inputs. ArticleMiner is abbreviated AM.}
\label{tab:grid_mltables}
\setlength{\tabcolsep}{6pt}
\begin{tabular}{llcccccc}
\toprule
& & \multicolumn{3}{c}{Raw PDF} & \multicolumn{3}{c}{Pre-parsed} \\
\cmidrule(lr){3-5}\cmidrule(lr){6-8}
Backbone & System & P & R & $F_1$ & P & R & $F_1$ \\
\midrule
Sonnet 4.6 & AM & 52.6 & 51.7 & 52.2 & 70.9 & 94.9 & 81.2 \\
 & Few-shot & 54.8 & 35.6 & 43.2 & 68.8 & 92.1 & 78.8 \\
Opus 4.6 & AM & 57.0 & 54.9 & \textbf{55.9} & 72.1 & 95.4 & 82.1 \\
 & Few-shot & 54.1 & 33.8 & 41.6 & 75.6 & 75.9 & 75.8 \\
Haiku 4.5 & AM & 45.7 & 59.4 & 51.6 & 70.8 & 92.8 & 80.3 \\
 & Few-shot & 52.3 & 29.9 & 38.0 & 79.9 & 75.4 & 77.6 \\
GPT-4o & AM & 55.0 & 38.9 & 45.6 & 78.2 & 92.4 & \textbf{84.7} \\
 & Few-shot & 41.8 & 27.0 & 32.8 & 80.9 & 91.5 & 85.9 \\
Gemini 2.5 Flash & AM & 44.7 & 40.8 & 42.7 & 63.9 & 92.2 & 75.4 \\
 & Few-shot & 44.5 & 40.5 & 42.4 & 75.5 & 99.0 & 85.7 \\
Qwen-2.5-7B (open) & AM & 49.2 & 51.3 & 50.2 & 75.3 & 79.3 & 77.3 \\
 & Few-shot & 38.7 & 13.5 & 20.0 & 76.4 & 59.7 & 67.0 \\
Mistral-7B (open) & AM & 42.8 & 21.5 & 28.6 & 58.5 & 63.7 & 61.0 \\
 & Few-shot & 36.9 & 12.7 & 18.9 & 53.6 & 37.3 & 44.0 \\
Llama-3.1-8B (open) & AM & 51.9 & 50.4 & 51.1 & 59.9 & 80.2 & 68.6 \\
 & Few-shot & 31.6 & 18.3 & 23.2 & 69.5 & 59.9 & 64.3 \\
\bottomrule
\end{tabular}
\end{table}

\begin{table}[!ht]
\centering\small
\caption{DiSCoMaT P/R/$F_1$ (\%). $n=111$ publications; raw-PDF and pre-parsed inputs. ArticleMiner is abbreviated AM.}
\label{tab:grid_discomat}
\setlength{\tabcolsep}{6pt}
\begin{tabular}{llcccccc}
\toprule
& & \multicolumn{3}{c}{Raw PDF} & \multicolumn{3}{c}{Pre-parsed} \\
\cmidrule(lr){3-5}\cmidrule(lr){6-8}
Backbone & System & P & R & $F_1$ & P & R & $F_1$ \\
\midrule
Sonnet 4.6 & AM & 98.2 & 66.4 & 79.3 & 91.5 & 91.1 & 91.3 \\
 & Few-shot & 82.7 & 52.9 & 64.5 & 92.1 & 94.7 & 93.4 \\
Opus 4.6 & AM & 97.0 & 69.5 & \textbf{81.0} & 95.1 & 94.2 & \textbf{94.7} \\
 & Few-shot & 80.4 & 47.2 & 59.5 & 91.2 & 85.5 & 88.3 \\
Haiku 4.5 & AM & 97.0 & 57.8 & 72.4 & 92.8 & 91.2 & 92.0 \\
 & Few-shot & 73.7 & 44.9 & 55.8 & 92.4 & 90.1 & 91.2 \\
GPT-4o & AM & 87.3 & 47.9 & 61.9 & 92.4 & 88.6 & 90.4 \\
 & Few-shot & 63.0 & 44.1 & 51.9 & 88.2 & 90.8 & 89.5 \\
Gemini 2.5 Flash & AM & 92.9 & 64.1 & 75.9 & 85.4 & 88.7 & 87.0 \\
 & Few-shot & 62.5 & 44.7 & 52.1 & 89.3 & 91.8 & 90.5 \\
Qwen-2.5-7B (open) & AM & 88.3 & 43.0 & 57.8 & 80.8 & 78.5 & 79.6 \\
 & Few-shot & 33.8 & 22.5 & 27.0 & 73.4 & 13.6 & 23.0 \\
Mistral-7B (open) & AM & 81.1 & 38.2 & 51.9 & 60.2 & 54.1 & 57.0 \\
 & Few-shot & 11.5 & 2.8 & 4.5 & 0.0 & 0.0 & 0.0 \\
Llama-3.1-8B (open) & AM & 79.0 & 10.3 & 18.2 & 69.3 & 73.8 & 71.5 \\
 & Few-shot & 15.4 & 26.0 & 19.4 & 60.0 & 26.0 & 36.3 \\
\midrule
Published GNN & & -- & -- & -- & -- & -- & 70.04 \\
\bottomrule
\end{tabular}
\end{table}

\begin{table}[!ht]
\centering\small
\caption{GeoChem P/R/$F_1$ (\%). Closed ArticleMiner runs use 28 entries and supplementary files; open runs use 26 entries. Few-shot reads PDF text. The reported P/R/$F_1$ aggregates need not satisfy the harmonic-mean identity across papers. ArticleMiner is abbreviated AM.}
\label{tab:grid_geochem}
\setlength{\tabcolsep}{6pt}
\begin{tabular}{llccc}
\toprule
Backbone & System & P & R & $F_1$ \\
\midrule
Sonnet 4.6 & AM & 58.0 & 82.9 & 60.8 \\
 & Few-shot & 3.0 & 0.5 & 0.8 \\
Opus 4.6 & AM & 58.1 & 83.5 & 60.6 \\
 & Few-shot & 8.0 & 2.1 & 3.0 \\
Haiku 4.5 & AM & 50.5 & 83.2 & 54.3 \\
 & Few-shot & 14.0 & 3.1 & 4.1 \\
GPT-4o & AM & 62.2 & 82.9 & \textbf{63.1} \\
 & Few-shot & 7.7 & 2.8 & 3.6 \\
Gemini 2.5 Flash & AM & 54.9 & 83.1 & 57.6 \\
 & Few-shot & 3.2 & 0.2 & 0.3 \\
Qwen-2.5-7B (open) & AM & 55.0 & 74.8 & 56.2 \\
 & Few-shot & 3.1 & 0.2 & 0.3 \\
Mistral-7B (open) & AM & 55.0 & 74.8 & 56.2 \\
 & Few-shot & 0.0 & 0.0 & 0.0 \\
Llama-3.1-8B (open) & AM & 55.0 & 74.8 & 56.2 \\
 & Few-shot & 4.6 & 1.9 & 2.6 \\
\bottomrule
\end{tabular}
\end{table}

\begin{table}[!ht]
\centering
\footnotesize
\caption{GPT-4o error distribution (\% of error events; complements the main-text Table~\ref{tab:error_dist}, which reports \textsc{ArticleMiner}-Sonnet and few-shot).}
\label{tab:error_gpt4o}
\setlength{\tabcolsep}{5pt}
\begin{tabular}{ll cccc c}
\toprule
\textbf{Task} & \textbf{System} & \textbf{Over-extraction} & \textbf{Omission} & \textbf{Attribute} & \textbf{Near miss} & $n_{\text{err}}$ \\
\midrule
\textsc{ChemTables} & \textsc{ArticleMiner}-GPT4o &  5.2 & 62.7 & 31.2 &  0.9 &  346 \\
\textsc{MLTables}   & \textsc{ArticleMiner}-GPT4o & 27.1 & 59.3 &  0.1 & 13.5 & 2020 \\
DiSCoMaT & \textsc{ArticleMiner}-GPT4o &  5.8 & 82.7 & 10.0 &  1.4 & 2788 \\
\bottomrule
\end{tabular}
\end{table}

\begin{figure}[!ht]
\centering
\begin{subfigure}[t]{0.48\linewidth}
\centering
\includegraphics[width=\linewidth]{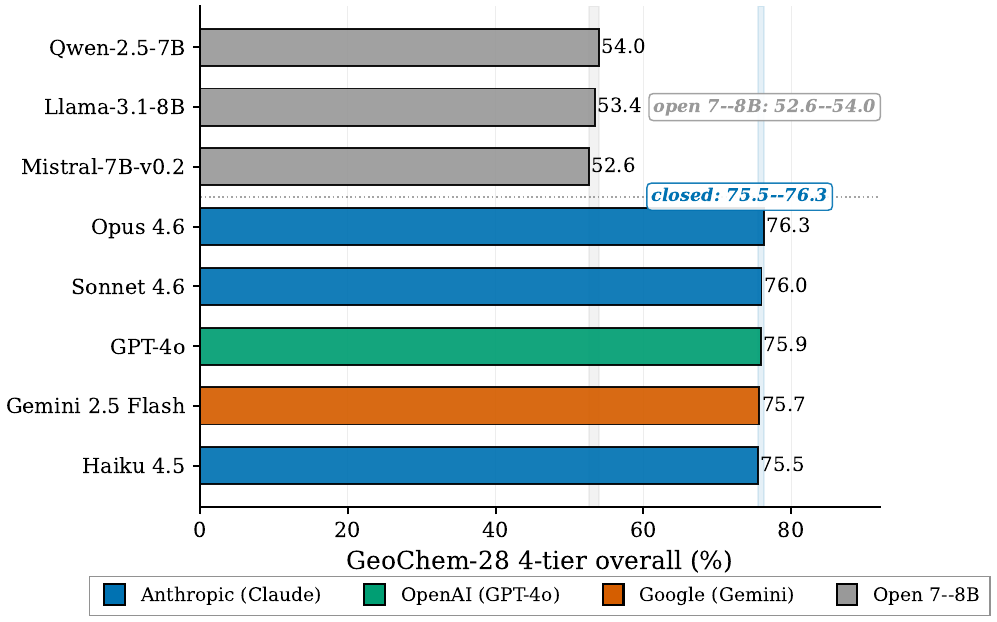}
\caption{Per-LLM four-tier Overall $S$ on \textsc{GeoChem}.}
\label{fig:cross_llm_invariance}
\end{subfigure}\hfill
\begin{subfigure}[t]{0.48\linewidth}
\centering
\includegraphics[width=\linewidth]{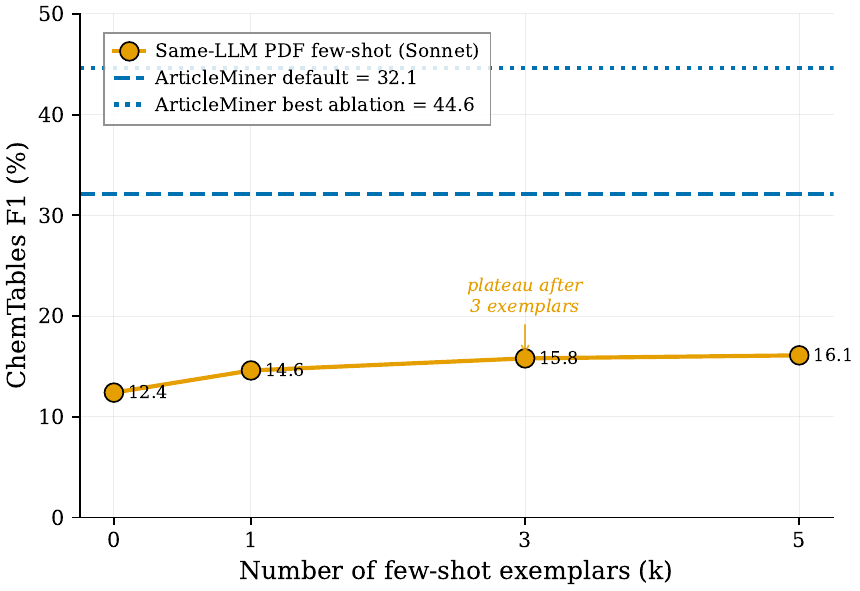}
\caption{\textsc{ChemTables} few-shot $F_1$ vs.\ $k$-shot exemplars (Sonnet~4.6).}
\label{fig:shot_scaling}
\end{subfigure}
\caption{Supporting analyses: per-LLM four-tier Overall \mbox{$S$} on \textsc{GeoChem} (left), and the few-shot exemplar-count plateau on \textsc{ChemTables} (right).}
\end{figure}

\begin{figure}[!ht]
\centering
\includegraphics[width=\linewidth]{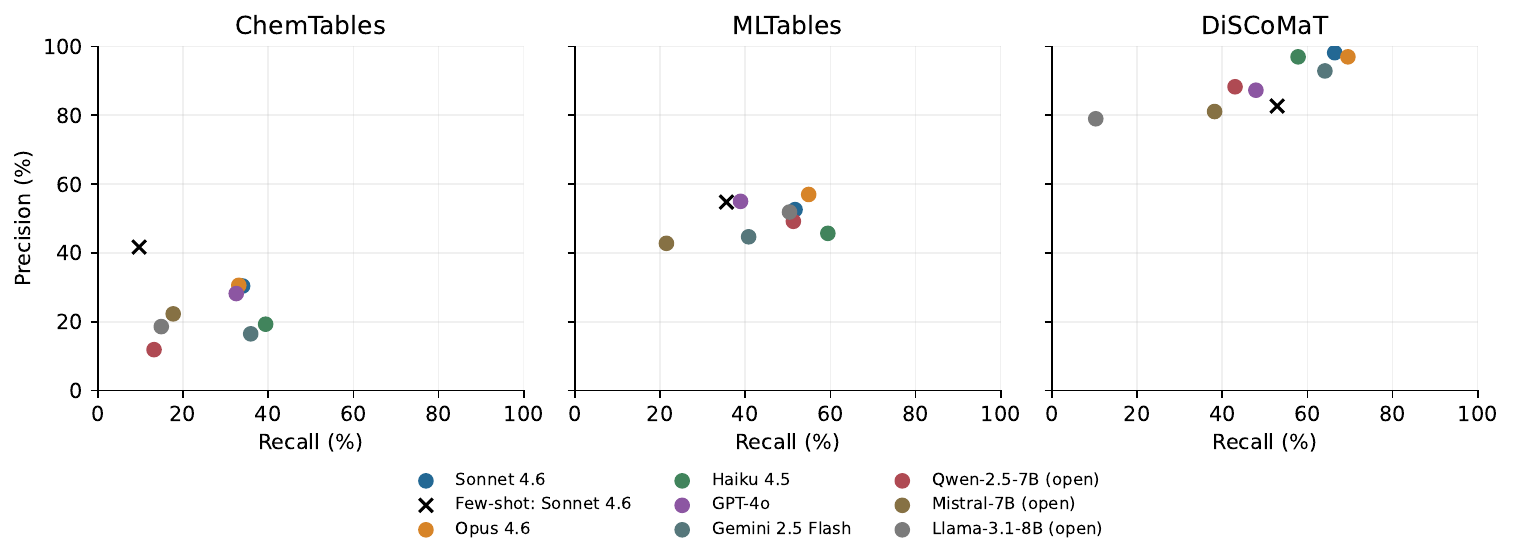}
\caption{Precision and recall on raw-PDF inputs, plotted directly from Tables~\ref{tab:consolidated_all}--\ref{tab:grid_discomat}. Circles denote ArticleMiner backbones; crosses denote the Sonnet few-shot baseline. Each panel uses its benchmark's strict tuple-matching metric.}
\label{fig:pr_scatter}
\end{figure}

\subsection{\textsc{GeoChem} per-paper 4-tier breakdown ($n=28$)}
Per-paper $T_1/T_2/T_3/T_4$ scores for the headline configuration (\textbf{Full (5-backend, \textsc{ArticleMiner})} from Table~\ref{tab:geochem_main}, Sonnet 4.6) yield the following bootstrap 95\% CIs on per-tier means ($n{=}28$, 1{,}000 resamples, seed 42): $T_1$ (metadata) $71.6\;[68.6,\,74.7]$, $T_2$ (numerical) $76.5\;[65.4,\,87.2]$, $T_3$ (structural) $68.5\;[55.0,\,80.4]$, $T_4$ (null) $90.6\;[85.2,\,95.1]$, \textbf{Overall} $76.0\;[70.1,\,81.4]$. The $T_2$ and $T_3$ intervals are wide ($\pm$10--13) because per-paper scores have a long left tail (a small number of layout-only or borderless-table papers score near 0); the overall-mean interval ($\pm$5.7) reflects heterogeneity across papers.

\subsection{Interpretation of component and backend analyses}
Table~\ref{tab:ablation}(a) reports component removals, whereas panel~(b) compares backend configurations. Their full/reference rows differ, so a backend effect must be computed against the corresponding all-five row in panel~(b). Removing Camelot changes ChemTables from 20.8 to 24.8, MLTables from 47.6 to 55.3, and DiSCoMaT from 71.0 to 72.5. GeoChem's backend comparison uses a 26-paper subset and a corrected ground-truth rerun, rather than the 28-paper headline reference. These distinctions prevent interpreting differences across panels as a single controlled ablation sequence.

\Needspace{14\baselineskip}
\subsection{Open-LLM \textsc{GeoChem} 4-tier breakdown ($n=26$, max-tokens=512)}
\label{app:open-geochem}
The three open 7--8B backbones run the full \textsc{ArticleMiner} pipeline on the 26-paper subset (the 2 data-reuse entries lacking standalone PDFs are excluded; see annotation protocol). Reported per-model four-tier scores:
\begin{itemize}[noitemsep,topsep=2pt,leftmargin=*]
\item Qwen-2.5-7B: $T_1{=}22.2$, $T_2{=}64.3$, $T_3{=}60.2$, $T_4{=}83.5$, Overall $=54.0$.
\item Mistral-7B-v0.2: $T_1{=}17.7$, $T_2{=}64.3$, $T_3{=}60.2$, $T_4{=}83.5$, Overall $=52.6$.
\item Llama-3.1-8B: $T_1{=}20.5$, $T_2{=}64.3$, $T_3{=}60.2$, $T_4{=}83.5$, Overall $=53.4$.
\end{itemize}
The reported aggregate $T_2/T_3/T_4$ scores are identical across these three runs, consistent with their shared supplementary-file parser. This does not imply identical cell-level outputs. All three deposit classifiers return malformed JSON under the 512-token cap, so the observed gap from closed models confounds model choice with decoding budget and coverage. Tables~\ref{tab:consolidated_all}--\ref{tab:grid_geochem} give the separately reported sample-score grid.

\section{Hardware, Software, and Reproducibility}
\label{app:repro}

Experiments used Ubuntu 22.04 and CUDA 12.4, with one NVIDIA H100 80GB GPU and vLLM 0.6.x for open-model inference. Closed models were called through the Anthropic, OpenAI, and Google SDKs in April--May 2026. Reported identifiers are listed below.
\begin{quote}\small\raggedright
\path{claude-sonnet-4-6}; \path{claude-opus-4-6};
\path{claude-haiku-4-5-20251001}; \path{gpt-4o-2024-08-06};
\path{gemini-2.5-flash}; \path{Qwen-2.5-7B-Instruct};
\path{Mistral-7B-Instruct-v0.2}; \path{Llama-3.1-8B-Instruct}.
\end{quote}
Reported parser versions are Docling 2.x, Marker 1.x, MinerU 0.9, pdfplumber 0.11, and Camelot 0.11. MinerU's OCR path was disabled when its weights were unavailable. The reported temperature is zero, and the resampling and exemplar-ordering seed is 42. The release contains stage-specific token caps and run configurations, but does not provide a complete manifest linking every historical result to its configuration. Table~\ref{tab:cost} reports per-paper cost and runtime.

The reported software versions are version families rather than a complete executable environment lock. The release also includes additional rotation handling and optional adaptive backend selection that were not part of the reported evaluated configurations. Reproducing those configurations requires the corresponding settings and output files, not merely running the latest default pipeline.

\section{Knowledge-Graph Output: Export Structure}
\label{app:jsonld}
Recovered samples are serialized as JSON-LD. The representation separates sample identity, elemental measurements, and source metadata, as described below.

\begin{description}[leftmargin=1.5em,style=nextline]
\item[\texttt{@context} and \texttt{@graph}.] The context declares a configurable default vocabulary and mineral/deposit namespace prefixes; the graph contains one node per sample row. A namespace declaration alone does not verify that the resulting terms correspond to a published ontology.
\item[Sample identity.] Each node has \texttt{@id}, \texttt{@type}, and \texttt{sample\_name}. The identifier incorporates the paper identifier and a normalized sample name. Optional fields include mineral, deposit, and analytical method.
\item[Measurements.] Each nonempty elemental value produces a measurement containing \texttt{element} and \texttt{value\_ppm}, with a unit when available. The exporter adds a below-detection note for the designated sentinel. A stated detection limit encoded as a negative value does not receive that note under the same rule; downstream consumers must therefore retain and interpret the source value convention.
\item[Source metadata.] The \texttt{provenance} object records the source PDF stem and pipeline model. These fields do not by themselves constitute a complete PROV mapping or cell-level evidence locator.
\end{description}

The task-level RDF binding describes the intended graph interface, while this exporter implements a particular representation. Native equivalent Turtle output, verified external ontology alignment, and downstream query correctness are not established by the tuple-level experiments. Such reuse requires validated predicate mappings, stable record identities, explicit censoring semantics, and compatible ingestion rules.

\section{\textsc{GeoChem}-28 Annotation Protocol}
\label{app:annotation}

This section summarizes the curation protocol used for the 28 contributed mineral-geochemistry papers.

\paragraph{Scope.} 28 peer-reviewed papers reporting LA-ICP-MS trace-element analyses of sulfide and sulphosalt minerals in ore deposits, published 2004--2025. Each paper is annotated against the 209-column \textsc{CMiO-MIN} schema
(an extension of CMiO/Hofstra~\cite{hofstra2021cmio} for individual-grain mineral geochemistry).

\paragraph{Annotators.} Multiple geochemistry annotators participated, coordinated with U.S.\ Geological Survey domain experts. Annotations were independently spot-checked by a second annotator and any disagreements reconciled against the source PDF + supplementary tables.

\paragraph{Per-paper procedure.}
\begin{enumerate}[noitemsep,topsep=2pt,leftmargin=*]
\item \textbf{Deposit metadata ($T_1$).} The fifteen annotated fields are \texttt{deposit\_name}, \texttt{deposit\_type}, \texttt{deposit\_environment}, \texttt{deposit\_group}, \texttt{all\_commodities}, \texttt{mineral}, \texttt{analytical\_method}, \texttt{instrument\_type\_model}, \texttt{laboratory\_location}, \texttt{operating\_conditions}, \texttt{standards\_used}, \texttt{country}, \texttt{age}, \texttt{host\_rock}, and \texttt{sample\_reference}. Values are transcribed from the publication where possible and otherwise normalized to the task vocabulary.
\item \textbf{Per-sample rows ($T_2$/$T_3$/$T_4$).} Each row corresponds to
one \texttt{(sample\_id, mineral)} analysis under a three-tier sample ID (top-level sample / sub-sample or thin section/spot). For each element column \texttt{<el>\_ppm} or \texttt{<el>\_wtpct}: a numeric value if reported; $-L$ for a stated detection limit $L$, or $-99999$ when below detection is reported without a limit; blank if not measured or not reported. \item \textbf{Mineral assignment.} One mineral per row. Mineral identity is taken from the per-analysis annotation in the paper's supplementary tables (\texttt{analysis\_id} prefix, data-sheet name, or explicit column), never inferred from abundance patterns.
\item \textbf{Units.} Preserved as reported; no conversion. Columns carry the unit in their name (e.g.\ \texttt{cu\_ppm}, \texttt{s\_wtpct}).
\end{enumerate}

\Needspace{6\baselineskip}
\paragraph{Conventions.}
\begin{itemize}[noitemsep,topsep=2pt,leftmargin=*]
\item \textbf{Below-detection limit:} $-L$ when the limit is known, or $-99999$ when it is unspecified. The encoding distinguishes censored measurements from missing values; the evaluator treats non-null gold entries in $T_2$ and gold-null entries in $T_4$.
\item \textbf{Not applicable / not measured:} blank cell.
\item \textbf{Reference materials} (MASS-1, NIST 610, etc.) are included as sample rows tagged with the reference-material flag. Annotation does not filter them out; both \textsc{ArticleMiner} and the few-shot baseline are expected to reproduce them in the prediction set so that the structural match in $T_3$ is well-defined.
\item \textbf{Data-reuse papers} (suffix \texttt{\_as\_reported\_in\_X\_et\_al\_YYYY}):
gold tuples are taken from paper~$X$'s supplementary tables. The original paper PDF is therefore not strictly required for evaluation, because all gold numbers reappear in the companion paper. Two of the 28 entries are data-reuse pairs; the open-LLM 4-tier breakdown in \S\ref{app:open-geochem} reports $n=26$ rather than $n=28$ because the open-LLM PDF-vision path requires standalone PDFs (not available for the two reuse entries), whereas the closed-LLM and headline runs report $n=28$ because they ingest the supplementary spreadsheets directly.
\end{itemize}

\paragraph{License.}
Ground-truth annotations are released under CC-BY-4.0. Source-paper PDFs retain their original licenses; the annotation license does not grant redistribution rights to those PDFs.


\end{document}